%% file: iclr2025_conference.tex
\documentclass{article} 
\usepackage{iclr2025_conference,times}

\input{math_commands.tex}

\usepackage{hyperref}
\usepackage{url}
\usepackage{algorithm}
\usepackage{algorithmic}
\usepackage{subfigure}
\usepackage{booktabs}
\usepackage{amsmath}
\usepackage{amssymb}
\usepackage{mathtools}
\usepackage{amsthm}
\usepackage{multirow}
\theoremstyle{plain}

\theoremstyle{definition}

\theoremstyle{remark}

\newcommand{\softargmax}{\textit{soft-argmax}}

\title{The Sampling-Gaussian for stereo matching}

\iclrfinalcopy

\author{Baiyu Pan \& Jichao Jiao \& Bowen Yao \& Jianxin Pang\\
Research Institute \\
Ubtech Robotics Corp\\
Shengzhen, China \\
\texttt{{baiyu.pan, jichao.jiao, bowen.yao, walton.pang}@ubtrobot.com}
\AND
Jun Cheng\\
Shenzhen Institute of Advanced Technology\\
Chinese Academy of Sciences\\
Shengzhen, China \\
\texttt{jun.cheng@siat.ac.cn}
}

\begin{document}

\maketitle
\input{src/chap0_abs.tex}
\input{src/chap1_intro.tex}

\input{src/chap2_pre.tex}

\input{src/chap3_observe.tex}

\input{src/chap4_method.tex}

\input{src/chap5_experiment.tex}
\input{src/chap6_conculsion.tex}

\bibliography{include/iclr2025_conference}
\bibliographystyle{iclr2025_conference}

\appendix
\input{src/chapF_appendix.tex}

\end{document}

%% file: math_commands.tex
\usepackage{amsmath,amsfonts,bm}

\def\eqref#1{equation~\ref{#1}}

\def\1{\bm{1}}

\DeclareMathAlphabet{\mathsfit}{\encodingdefault}{\sfdefault}{m}{sl}
\SetMathAlphabet{\mathsfit}{bold}{\encodingdefault}{\sfdefault}{bx}{n}



%% file: src/chap0_abs.tex
\begin{abstract}





The \textit{soft-argmax} operation is widely adopted in neural network-based stereo matching methods to enable differentiable regression of disparity. However, network trained with \textit{soft-argmax} is prone to being multimodal due to absence of explicit constraint to the shape of the probability distribution. Previous methods leverages Laplacian distribution and cross-entropy for training but failed to effectively improve the accuracy and even compromises the efficiency of the network. In this paper, we conduct a detailed analysis of the previous distribution-based methods and propose a novel supervision method for stereo matching, \textit{Sampling-Gaussian}. We sample from the Gaussian distribution for supervision. Moreover, we interpret the training as minimizing the distance in vector space and propose a combined loss of L1 loss and cosine similarity loss. Additionally, we leveraged bilinear interpolation to upsample the cost volume. Our method can be directly applied to any \textit{soft-argmax}-based stereo matching method without a reduction in efficiency. We have conducted comprehensive experiments to demonstrate the superior performance of our \textit{Sampling-Gaussian}. The experimental results prove that we have achieved better accuracy on five baseline methods and two datasets. Our method is easy to implement, and the code is available online.

\end{abstract}

%% file: src/chap1_intro.tex
\section{Introduction}
\input{src/figandtable/fig1.tex}
\label{section1}
Stereo matching is a fundamental topic of computer vision which has been a subject of extensive research for many years. 
Accurate stereo matching is essential for deriving scene depth through the triangulation by the displacement of corresponding points in binocular images. The applications of stereo matching cover a wide range of advanced technologies, including autonomous driving, robot navigation, and drone control.

The common baseline of end-to-end learning-based stereo matching \citep{mayer_large_2016} comprises three key modules: feature extraction, cost volume aggregation, and \textit{soft-argmax}-based disparity regression \citep{GC-net}. Features are extracted from the input image pair via a siamese network architecture. Subsequently, a 5D ($B,C,D,H,W$) cost volume is generated by concatenating features from the left and right images, while the disparity is the extra dimension $D$. This cost volume then serves as input to a disparity regression module, which employs 3D convolutions to refine the output. \cite{GC-net} was the first to leverage \textit{soft-argmax} to achieve differentiable regression of disparity. Its efficiency and simplicity have made it a popular baseline for numerous subsequent studies \citep{chang_pyramid_2018,pan_multi-stage_2020,wang2021pvstereo,ACVnet,shen2023digging}. 
In the pursuit of accuracy, various innovative modules have been proposed for improvement, feature fusion\citep{xu_aanet_2020,guo_group-wise_2019}, robust aggregation\citep{GA_net,shamsafar_mobilestereonet_2021}, iterative regression\citep{teed2021raft,xu2023iterative, xu2024igevpp}. Nevertheless, \textit{soft-argmax} remains a key component of these methods.



As the cost volume went through 3D CNNs, the channel is progressively reduced to 1. Then \textit{soft-argmax} (Equ. \ref{softargmax}) module is applied to obtain the disparity. 
\begin{equation}
    d =  \sum_{i} i*softmax(z_{i}) =\sum_{i} i*\frac{e^{z_{i}}}{\sum e^{z_{i}}}.
\label{softargmax}
\end{equation}
$d$ denotes the averaged disparity. $i$ and $softmax(z_{i})$ denotes the index of disparity and the probability of $i$. 
\begin{equation}
smoothl1(d,\hat{d}) = \left\{\begin{matrix}
    0.5(d-\hat{d})^{2},& if |d-\hat{d}|<1\\
    |d-\hat{d}|-0.5,& otherwise
    \end{matrix}\right.,
    \label{smoothl1}
\end{equation}
Then a smoothl1 loss (Equ. \ref{smoothl1}) is used to measure the distance between $d$ and ground-truth $\hat{d}$.
As \textit{soft-argmax} is widely adopted, researchers have also noticed its limitations. \cite{GC-net} regarded the \textit{soft-argmax} as a probability distribution of disparity and point out it's prone to being influenced by multimodal distribution as it estimates a weighted summation of all modes. Similarly, \cite{Chen_2019_ICCV} demonstrated that the averaged disparity of multimodal is deviated from the center of the dominating mode. They conclude that the ambiguous matching is the cause of multimodal problem. Researchers have proposed various methods that aimed to solve the problem \citep{ICRA_similar, bangunharcana2021correlate, PDS, Xu_2024_CVPR}. Those methods can be broadly summarized as two steps, constructing a direct supervision signal for the probability distributions to be predominately unimodal, and limiting the disparity range of \textit{soft-argmax} through a post-processing.

It's challenge to reduce the ambiguous matching relies on network's regularization only. Therefore, \cite{PDS} taken the ground-truth disparity as the center $\mu$ of a discrete Laplacian distribution, 
\begin{equation}
    q(x)=\frac{1}{2b}e^{\frac{-|x-\mu|}{b}},
\label{Laplacian}
\end{equation}
where $q(x)$ is the probability of integer $x$.
And they optimize the network with cross-entropy loss,
\begin{equation}
    H(p,q) = \sum_{x\in [d_{min},d_{max})}p(x) log(q(x)),
\label{cross-entropy}
\end{equation}
$p$ is the estimated probability.
Following their ideas, different distribution are adopted, Gaussian\citep{Chen_2019_ICCV}, Laplacian\citep{PDS,Xu_2024_CVPR, LocalSimilarity, AdaptiveUnimodal} and Dirac impulse \cite{ICRA_similar}, etc. This first step will effectively force the network  learns to estimate a distribution that centered at the highest likelihood. To further mitigate the effects of full-band weighted summation, most of the methods proposed limiting the disparity range of the summation to the neighbors of the highest likelihood. 
Those methods have two issues. Firstly, the network trained with cross-entropy tends to locate the highest likelihood. However, due to the absence of explicit constraints, the values of each distribution are imprecise, resulting in a deviated disparity. The second step is implemented through Top-k or equivalent operation. But consequently results in an efficiency reduction due to the operation is not parallelizable.

To address those problems, we propose a novel Gaussian distribution-based supervision method with combined loss for stereo matching called \textit{Sampling-Gaussian}. As shown in Fig. \ref{fig1}, our method achieves notable improvement over the listed commonly used baselines. Additionally, our method does not rely on Top-k or any other post-processing. Our method can be directly applied to any \textit{soft-argmax}-based stereo matching algorithms without a decrease in efficiency. 
In section 3, we conduct a theoretical analysis of \textit{soft-argmax} to fundamentally explain the reason of distribution-based supervision outperforms single value-based supervision. 
Moreover, we explore why previous methods have failed to improve the accuracy directly and concluded two reasons. First is the settings of the minimum and maximum disparity, which was empirically set to $0$ and $192$. Consequently, the regression near the endpoints are overlooked. Second is the trilinear interpolation which was used to upsample the cost volumes. The upsampled possibility distribution is impossible to fit the target distribution. We have conducted comprehensive ablation studies to illustrate the necessity of our proposed modules. Furthermore, we have implemented our method with five popular baselines\citep{chang_pyramid_2018,shamsafar_mobilestereonet_2021,guo_group-wise_2019,xu2023iterative} to demonstrate that our method is easy to implement and universally applicable. At last, our method has also achieved state-of-the-arts results on Sceneflow\citep{Mayer2016CVPR} and Kitti2012, \citep{kitti2012}, Kitti2015\citep{kitti2015}.

In conclusion, our contributions has three folds:
\begin{itemize}
\item Our proposed \textit{Sampling-Gaussian} can effectively improve the accuracy of stereo matching method. Moreover, we provide a new view by our interpretation of the distribution-based training.

\item We have identified the two fundamental reasons that lead to the inferior performance of distribution-based methods. Moreover, we provide theoretical explanations and solutions.

\item We conducted comprehensive experiments to demonstrate that \textit{Sampling-Gaussian} can be implemented to various methods and achieves notable improvement. Our method is easy to implement and code is open-sourced.
\end{itemize}


%% file: src/figandtable/fig1.tex
\begin{figure}[h]
    \centering
    \includegraphics[width=0.49\linewidth]{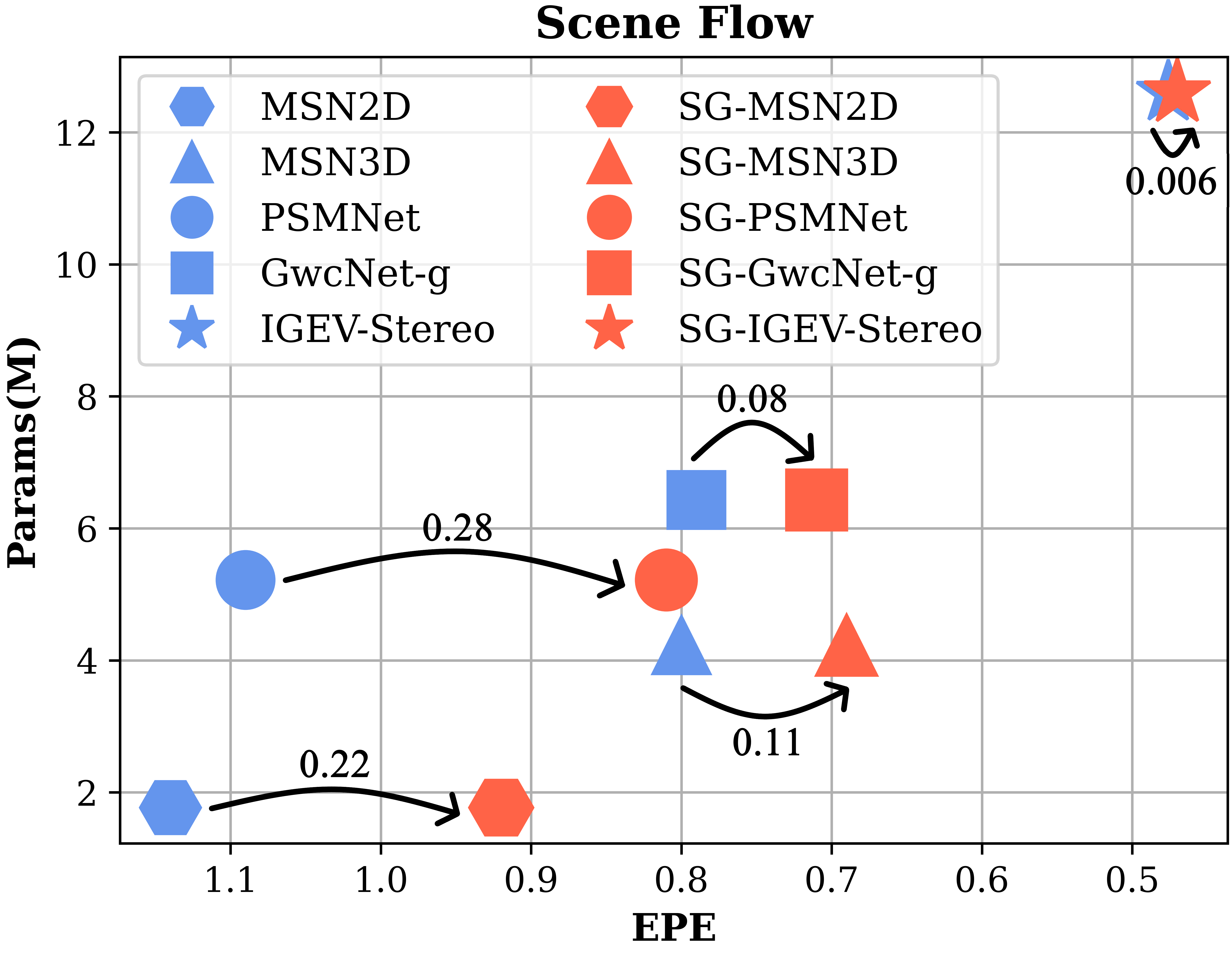}
    \includegraphics[width=0.49\linewidth]{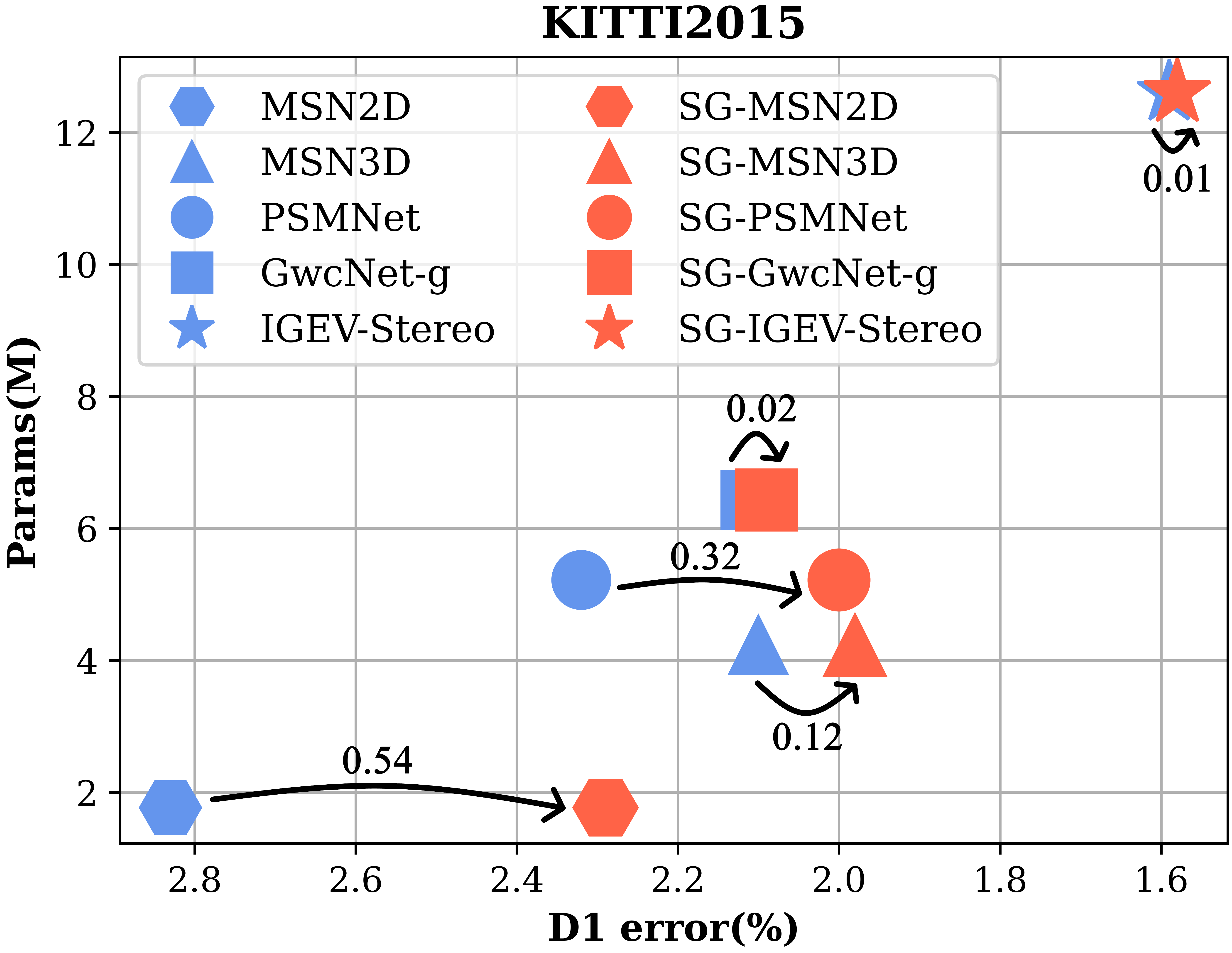}
    \caption{Quantitative comparisons on Sceneflow and Kitti. We implement our \textit{Sampling-Gaussian} (SG) with five baseline methods for comparison. They are MSN2D and MSN3D \citep{shamsafar_mobilestereonet_2021}, PSMnet\citep{chang_pyramid_2018}, GwcNet-g\citep{guo_group-wise_2019}, IGEV-Stereo\citep{xu2023iterative}}
    \label{fig1}
\end{figure}

%% file: src/chap2_pre.tex
\section{Related Works}

\subsection{The baseline of stereo matching}
Stereo matching method is a method that calculates the disparity map of the binocular images with size $(H,W)$. The feature network\citep{GC-net,kaiming_resnet_2015} extracts the features with size $(\frac{H}{4}, \frac{W}{4})$. Then the cost volume is constructed with size $(\frac{d_{max}}{4},\frac{H}{4},\frac{W}{4})$, where $d_{max}$ is a hyperparameter and empirical set to 192\citep{chang_pyramid_2018}. A disparity regression network with 3D convolutions is used for refine the cost volume. Its output remains the same size as cost volume. Then a trilinear interpolation operation is used for upsample the output to $(d_{max}, H, W)$. At last, a \textit{soft-argmax} operation is applied.

\subsection{Improvement methods}
Based on the baseline, the subsequent proposed improvement methods can be classified into several levels: feature level, module level, baseline level, and distribution level. Firstly, at the feature level, \cite{chang_pyramid_2018} proposed the PSMNet which adopts a spatial feature pyramid\citep{SPP_He} to extract and fuse multi-resolution features, and stacked-hourglass module is adopted as regression module to improve the refinement. Based on PSMNet, \cite{guo_group-wise_2019} proposed a group-wise correlation network(GwcNet) which calculates the dot products of the left and right features instead of a concatenation. And at module level, \cite{GA_net} proposed a guided-aggregation module to better refine the cost volume. And \cite{ACVnet} leverages attention mechanism to supervise the cost volume. At the baseline level, researchers proposed new baselines to improve the accuracy of the efficiency. \cite{xu_aanet_2020} and \cite{pan_multi-stage_2020} proposed to progressively aggregate the cost volume to the full size. Others proposed 2DConv-based methods\citep{pan_2024_icra,shamsafar_mobilestereonet_2021} to reduce the high FLOPs. 
And \cite{xu2023iterative} proposed to iterative refine the disparity and significantly improve the accuracy but at the expense of speed.






\subsection{Distribution-based improvement method}

The \textit{soft-argmax} operation is widely applied in various tasks as it retrieves the index of the highest probability in a differentiable way. 
Despite its efficiency, researchers continuously explore and propose better methods. From a distribution-view, the \textit{soft-argmax} is equivalent to retrieves the mean of the probability distribution\citep{li2021localization}. Consequently, network trained with \textit{soft-argmax} lacks explicit supervision for the distribution, resulting in unconstrained probability shape.

However, this disadvantage of \textit{soft-argmax} receives less attention than other aspect. Because, as the network become deeper and larger, the multimodal problem can be solved partially by the network's generalization ability. 
The DSNT \citep{DSNT} introduced a differentiable operation to render the heatmap with a 2D Gaussian kernel as a constraint for shape. 
Some methods attribute inaccurate estimates to the multimodal problem. The PDS \citep{PDS} limit the range of the \textit{soft-argmax} with Top-k during inference in order to solve the multimodal problem. An unresolved issue with PDS is its lack of robustness, as the range parameter is set in advance. A corresponding solution was proposed in \cite{ICPR_softargmax}, using learned weights to suppress unreliable disparity regions. A similar idea was proposed in \cite{ICRA_similar}, where they use a \textit{Dirac impulse} to model the distributions.

%% file: src/chap3_observe.tex
\section{Explorations}
In this section, we first analyze the biased gradient of \textit{soft-argmax} to establish that distribution-based supervision is necessary for stereo matching. Then, we analyze the two basic settings that have led previous distribution-based methods to their inferior improvements.



\subsection{Analysis of biased gradient}
We first analysis the partial differential equation of \softargmax.   
The $e^{z_{i}}$ denotes the input of \textit{softmax}. The partial derivative of \textit{$e^{z_{i}}$} is defined as
\begin{equation}
\begin{aligned}
\frac{\partial L}{\partial e^{z_{i}}}
&=\frac{\partial L}{\partial d}\frac{\partial d}{\partial e^{z_{i}}}\\
&=
\frac{\partial L}{\partial d}(i\frac{e^{z_{i}}}{\sum e^{*}}(1-\frac{e^{z_{i}}}{\sum e^{*}})
+\sum_{j\neq i}j(-\frac{ e^{z_{j}}}{\sum e^{*}}*\frac{e^{z_{i}}}{\sum e^{*}}))\\
&=\frac{\partial L}{\partial d}(\frac{ e^{z_{i}}}{\sum e^{*}}({ i}-d)).
\end{aligned}
\label{softargmaxequ}
\end{equation}
The variable $i$ denotes the corresponding index, and $d$ denotes the result of Equation \ref{softargmax}. It is evident that $i$ would always receive a weight $(i-d)$ to the gradient that is proportional to its distance to $d$. Therefore, it is difficult for the network to reach the global minimum since the gradient is biased.






\subsection{Analysis of distribution-based method}
Two basic settings are widely adopted, the disparity range $[0,192)$, and trilinear interpolation.

\begin{figure}
    \centering
    \includegraphics[width=0.4\linewidth]{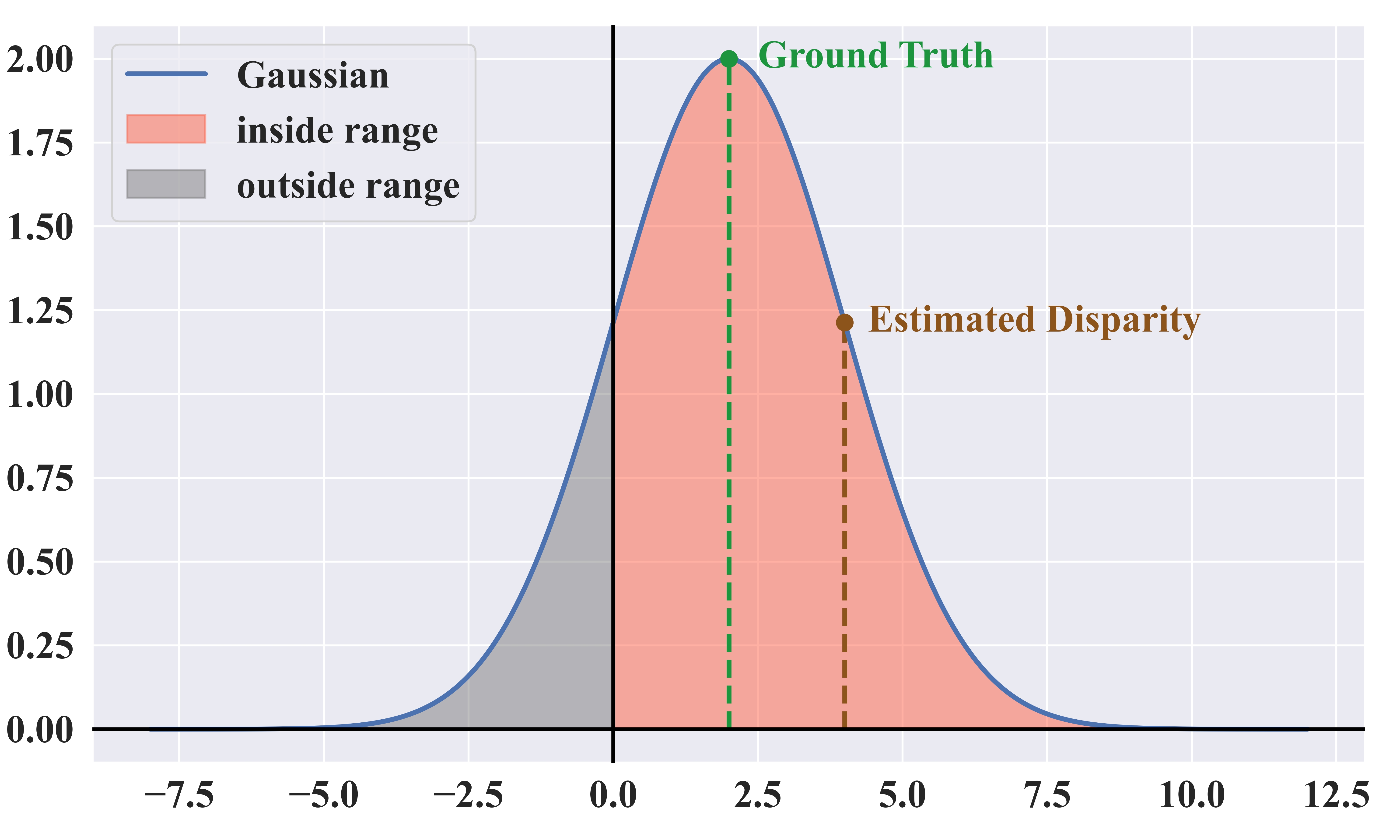}
    \includegraphics[width=0.4\linewidth]{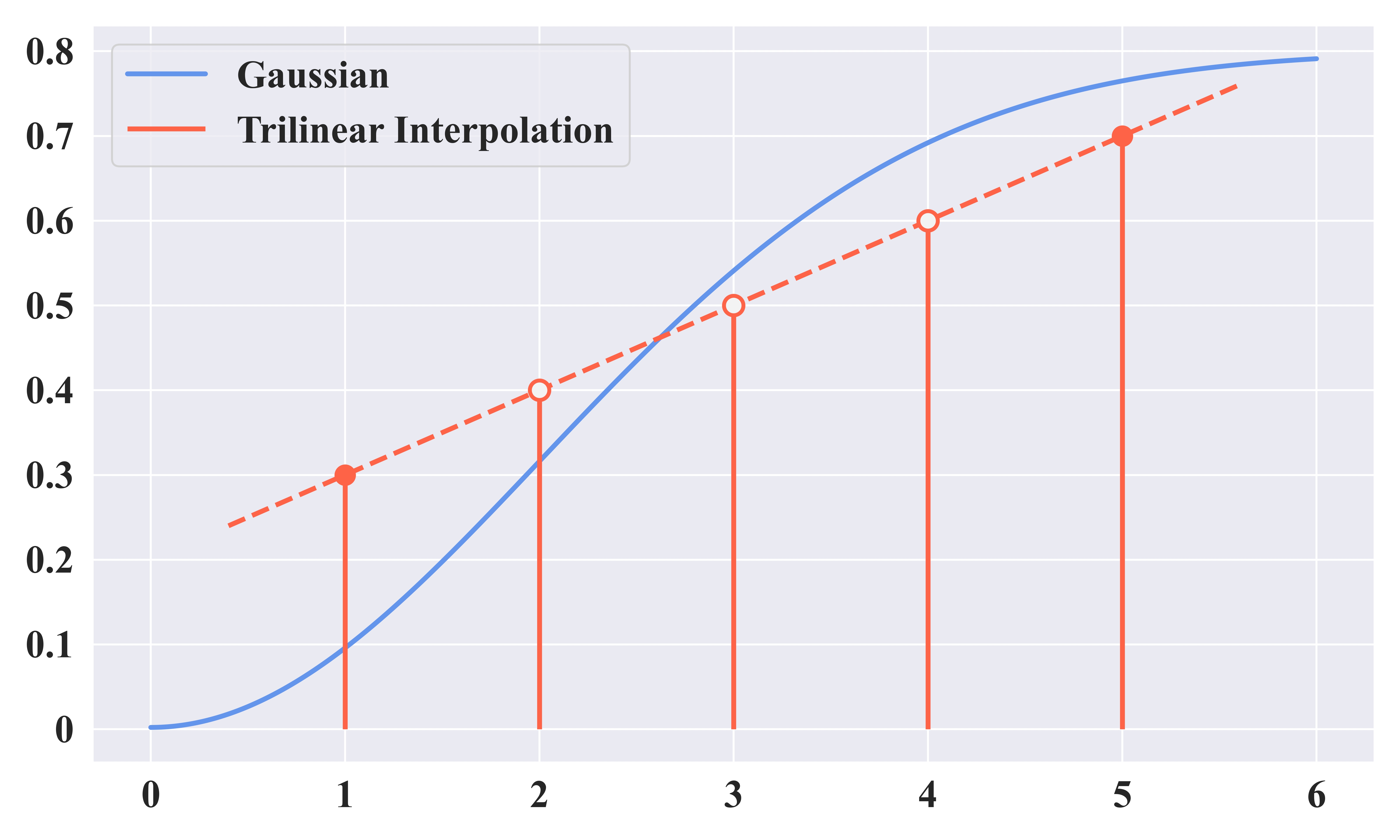}
    \caption{ The \textit{left} plot shows the impact of the disparity range. The predicted disparity near the
    endpoints is deviated. The \textit{right} plot shows that after trilinear interpolation, the probabilities are
    linearly distributed and are unable to fit the Gaussian distribution.}
\end{figure}

a) This setting of range is inherited from the \textit{soft-argmax}-based method. However, for distribution-based method, it would lead to a deviated disparity. For instance, a distribution $q$ is sampled based on Equ (\ref{Laplacian}) when ground-truth equals $0$. And the expectation of q, which is equivalent to calculates the \textit{soft-argmax}, is $0$ when the range is infinite. 
\begin{equation}
    \sum_{x=-\infty}^{\infty} x*q(x) =0 < \sum_{x=0}^{\infty} x*q(x).
\end{equation}
And if the minimum disparity is set to $0$, the expectation of q would be deviated. It's same for the maximum disparity.

b) After the trilinear interpolation, the cost volume is linearly resized from $\frac{d_{max}}{4}$ to $d_{max}$. However, whether the Laplacian or Gaussian distribution are taken for the supervision, their distribution is convex. As a result, it is impossible for the network to converge.

%% file: src/chap4_method.tex
\section{The proposed \textit{Sampling-Gaussian}}

\input{src/figandtable/workflow.tex}

In this paper, we present an innovative interpretation of the \textit{soft-argmax} and distribution-based supervision from the perspective of vector space. Therefore, the training process can be regarded as minimizing the distance between two vectors. Based on this interpretation, we propose the \textit{Sampling-Gaussian}, which consists of three parts.

\subsection{Construct the distribution}
First, we leveraged the probability density function of Gaussian distribution to sample the discrete supervision signal within an extended disparity range. The original disparity range is $[0,d_{max})$, we extend the range to $D=[-d_{ext},d_{max}+d_{ext})$, the size of D$=d_{max}+2*d_{ext}$. The sampling function is defined as
\begin{equation}
    q(x)=\frac{e^{-\frac{(x-\mu)^2}{2\sigma^{2}}}}{\sum_{x} e^{-\frac{(x-\mu)^2x-\mu}{2\sigma^{2}}}}.
    \label{sampling-gaussian}
\end{equation}
The $\mu$ is the ground-truth disparity.
$\sigma$ is used to control the shape, and $0.5$ achieves the best result.

\subsection{Structure alterations}
As we extended the disparity range by $d_{ext}$, the size of cost volume $C$ is also changed. The construction of $C$ involves iteratively constructing the $C$ by shifting the feature map by $1$ pixel,
\begin{equation}
C(c,d,x,y)=g(f_{l}(.,x,y),f_{y}(.,x-d,y)).
\end{equation}
The $f_{l},f_{r}$ denotes the features of left and right image. And $g$ denotes a fusion method for features, usually is group-wise correlation\citep{guo_group-wise_2019} or concatenation\citep{chang_pyramid_2018}. And the size of $C$ is $[B, C, \frac{D}{4}, \frac{H}{4},\frac{W}{4}]$.

Then, a \textit{bilinear interpolation} is leveraged to upsample the cost volume after the regression modules,
\begin{equation}
    \textbf{C} = bilinear(C).
\end{equation}
And the size of $\textbf{C}$ is $[B,C,\frac{D}{4}, H, W]$.

\begin{figure}
    \centering
    \includegraphics[width=0.32\linewidth]{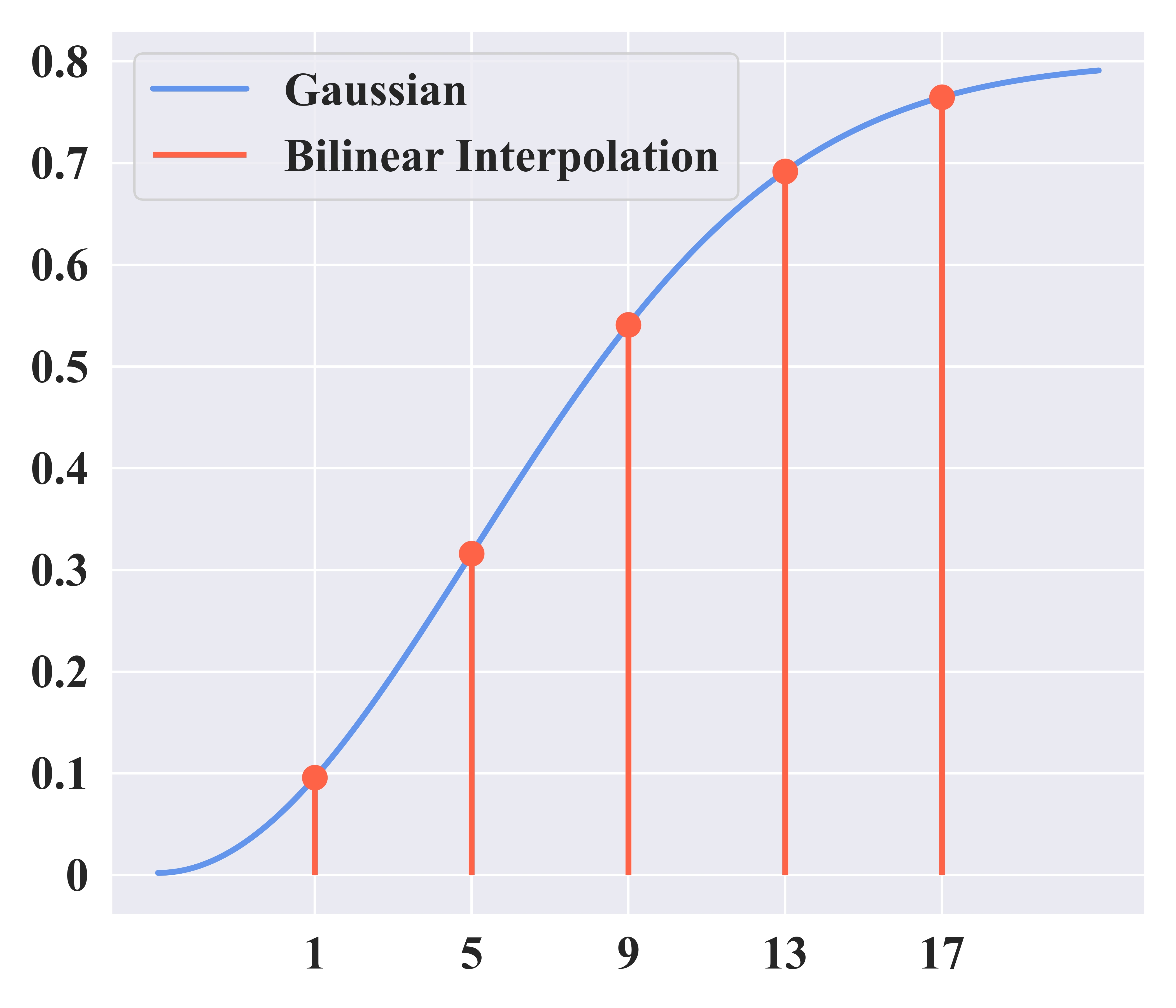}
    \includegraphics[width=0.32\linewidth]{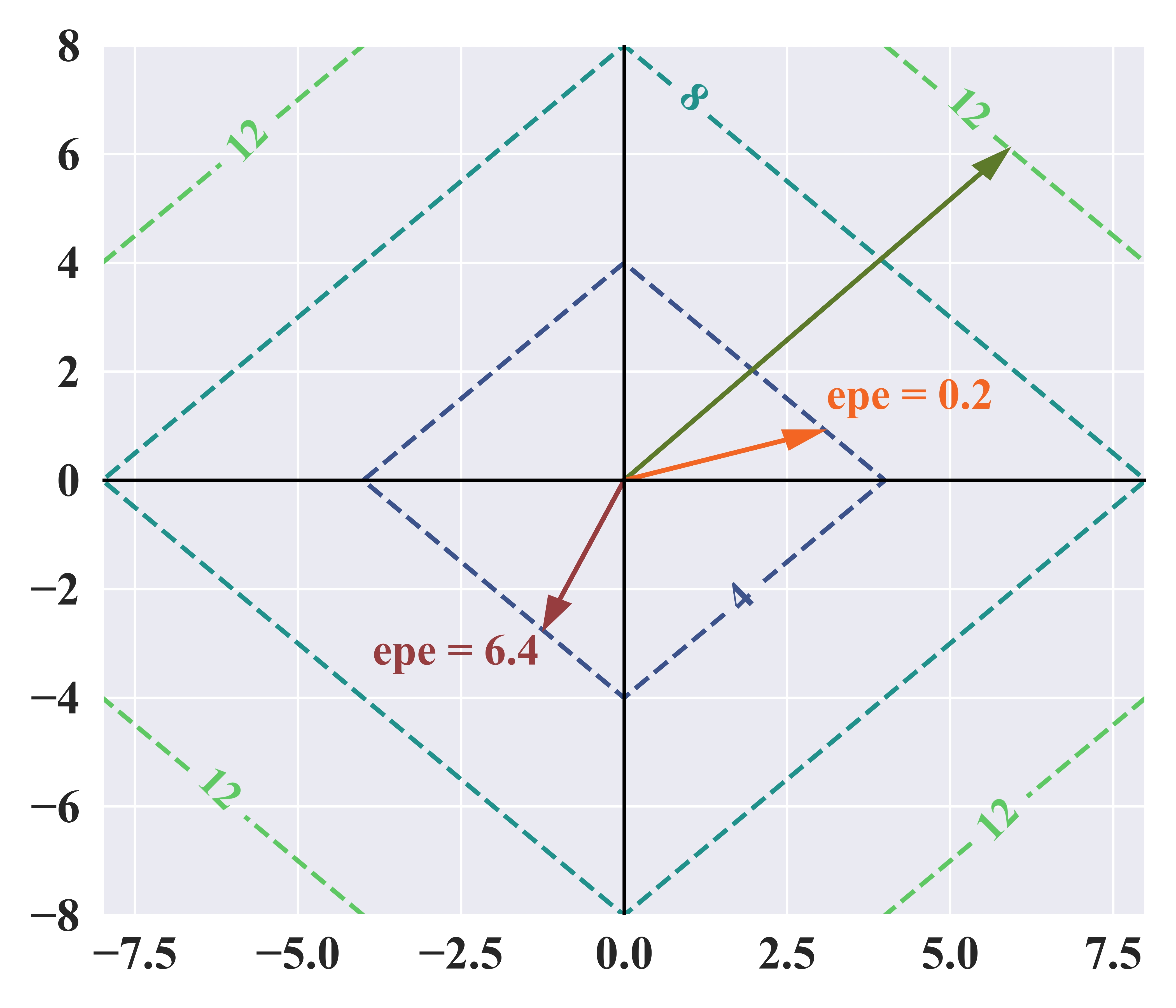}
    \includegraphics[width=0.32\linewidth]{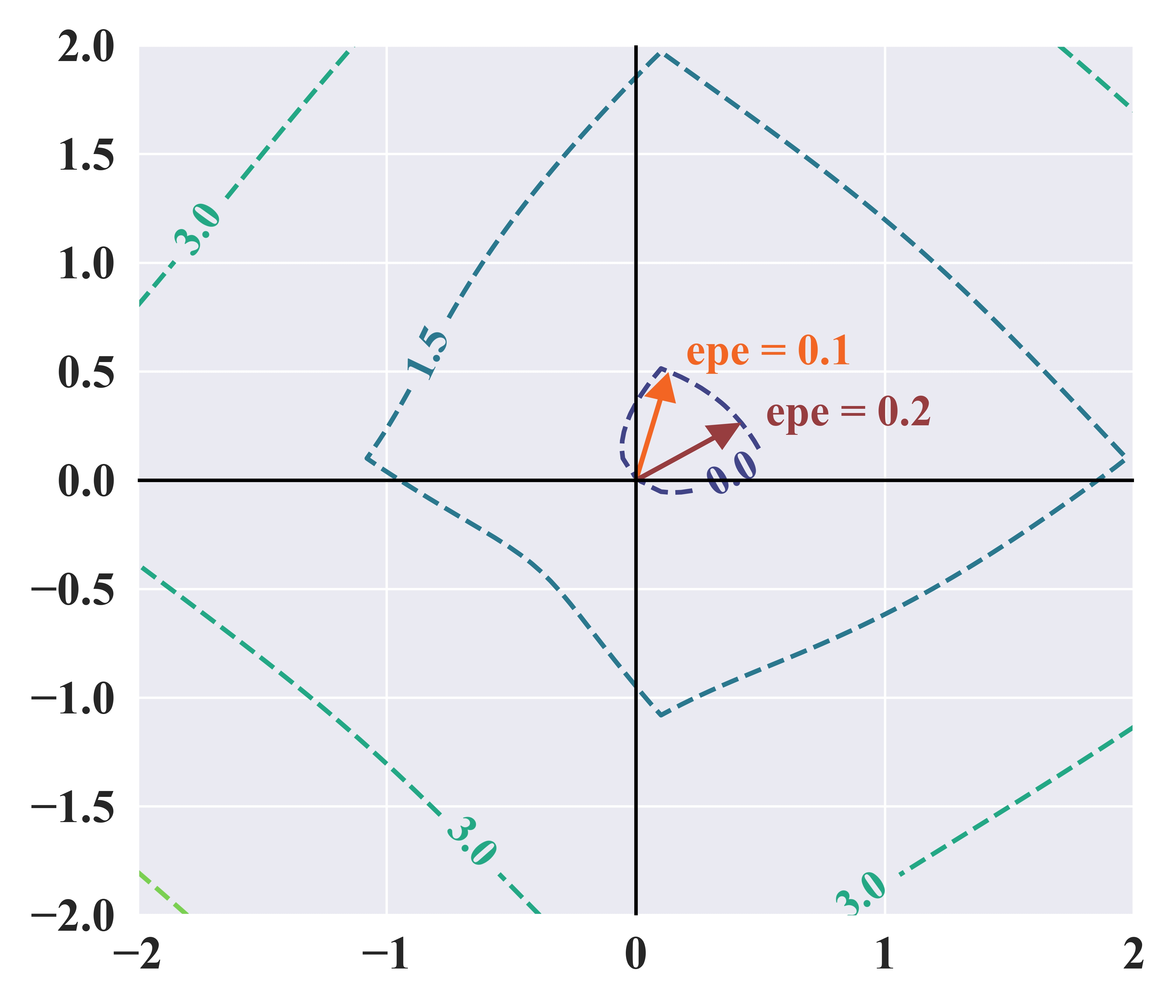}
    \caption{The \textit{left} plot: Since the probabilities are not linearly related, therefore it can fit into the Gaussian distribution. The \textit{middle} plot: The loss landscape of L1 loss, dashed lines are contour line. Two vectors with the same L1 loss, could result in significant difference in EPE. The \textit{right} plot: The loss landscape of the combined loss and the EPE of two vectors with the same loss are similar.}
    \label{aba_fig3}
\end{figure}

\subsection{Combination loss}
The supervision distribution is $q$. The output of network is distribution $p=softmax(\textbf{C})$. The cross-entropy\citep{PDS,Chen_2019_ICCV,Xu_2024_CVPR} loss can constrain the majority parts of the distribution but failed to optimize the distribution to further to exact value. As we interpret $p$ and $q$ as two vectors, we leverage L1 loss to measure the distance, 
\begin{equation}
    L_{1}(p,q) = \frac{1}{n}\sum_{i}^{n}|p(i)-q(i)|,
\end{equation}
L1 loss is sensitive to all difference of value regardless of the index. However, this is also the disadvantage. As depicted in Fig. \ref{aba_fig3}, that two vectors with same L1 loss to $q$, their predicted disparity could be very different.

In repose, we have proposed a negative cosine similarity to measure the difference in direction between $p$ and $q$,
\begin{equation}
    L_{cos}(p,q)=-\frac{\sum_{i}^{n} p(i)q(i)}{\sqrt{\sum p(*)^{2}}\sqrt{\sum q(*)^{2}}}.
\end{equation}
Then the two losses is combined with a weight $\lambda=0.5$,
\begin{equation}
    L(p,q) = L_{1}(p,q) + \lambda* L_{cos}(p,q).
\end{equation}
As shown in Fig. \ref{aba_fig3}, the vectors on the same contour line has similar end-point error(EPE).

\subsection{Inference}
A key contribution of our method, is that we do not rely on Top-k operation for refinement.
During the inference, we calculate the expectation of $p$ directly,
\begin{equation}
d = 4*\sum i*p = 4*\sum_{D/4} i* softmax(\textbf{C}_{i}),
\end{equation}
which has the same form of \textit{soft-argmax}. Therefore, our method can be easily implemented with most of the \textit{soft-argmax}-based method. After the upsampling, the dimension of disparity remains one-fourth of original. Consequently, the value after regression is also one-fourth of the original. Thus, the ``$4*$'' is to recover the disparity to its original scale.



%% file: src/figandtable/workflow.tex
\begin{figure}[htbp]
    \centering
    \includegraphics[width=0.99\linewidth]{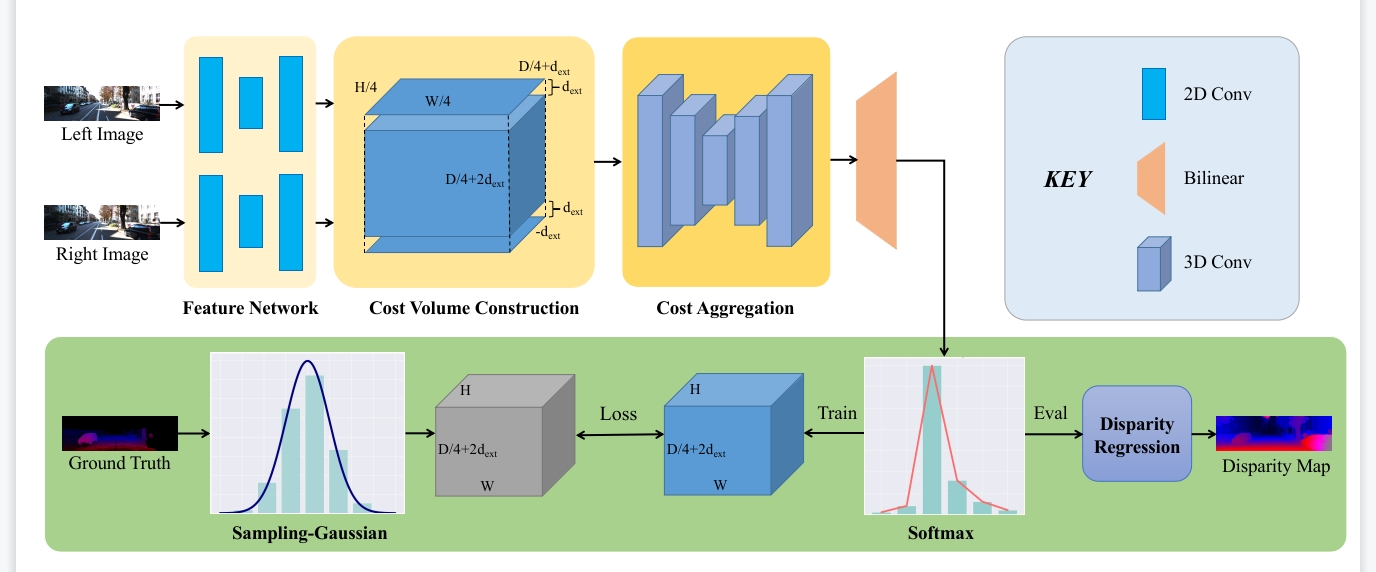}
    \caption{The workflow of our proposed \textit{Sampling-Gaussian}}
\end{figure}

%% file: src/chap5_experiment.tex
\section{Experimental results}
In this section, we report our implementation details and experimental results. We have implemented \textit{Sampling-Gaussian} with $5$ most representative methods for comparisons:


1. \textit{PSMNet}\citep{chang_pyramid_2018}. The ``ResNet'' of the stereo matching. They outperformed SOTA algorithm by $5\%$ at the time. Their method is open-source, easy to read and replicate. We use this method for a wider range of comparisons.

2. \textit{GwcNet-g}\citep{guo_group-wise_2019}.  A group-wise correlation module is proposed based on PSMNet. Their module is widely adopted. Code is open-sourced.

3$\&$4. \textit{MSN3D} and \textit{MSN2D} \citep{shamsafar_mobilestereonet_2021}: They have proposed lightweight networks by leveraging 2D convolutions to reduce computational expenses while maintaining accuracy. 

5. \textit{IGEV-Stereo}\citep{xu2023iterative}:
Based on RAFT\citep{teed2021raft}, they proposed an iterative refine module and achieves SOTA results. We implement our method with IGEV-stereo to demonstrates our methods is compatible with a variety of structure.



We conducted experiments mainly on two \textbf{datasets}:
\textbf{Sceneflow}\citep{mayer_large_2016} is a large scale of synthetic stereo dataset which contains more than 35k training pairs and 4.3k testing pairs with resolution 960x540. 
\textbf{Kitti}\citep{kitti2012,kitti2015} We use \textit{Kitti2012} and \textit{Kitti2015} for train and test. They contain 395 pairs for training and 395 pairs for testing in total, with resolution $1242\times 375$.

\subsection{Implementation details}
For simplicity, we will refer to our \textit{Sampling-Gaussian} as SG. 
Our implemented versions of method are denoted as SG-PSMNet or SG-MS2D.
Our method is implemented using the PyTorch framework. We conducted all the experiments on two A100 GPUs.
We leverage AdamW\citep{Adamw} with $\beta_{1}=0.9,\beta_{2}=0.999$, weight decay$=10^{-2}$, as optimizer. All the networks are trained with similar protocol: train on Sceneflow for $20$ epochs with learning rate$=10^{-3}$. Then, finetuning on Kitti for $200$ epochs with lr$=10^{-3}$, then with lr$=10^{-4}$ for another $300$ epochs, and with lr$=10^{-5}$ for the last $300$ epochs. For IGEV-stereo and MSN2D, the parameters are slightly changed.
Two metrics are adopted for evaluation (both are lower the better):  
\textit{End-point error} (EPE)\citep{mayer_large_2016}, commonly used in optical flow. It calculates the l1 loss. 
\textit{D1 error} \citep{kitti2015} calculates the percentage of error pixels. Pixels with EPE larger than 3 will be considered as error.





\subsection{Ablation studies}

\subsubsection{Sigma $\sigma$ of the \textit{Sampling-Gaussian}}
The $\sigma$ controls the shape of the distribution and directly affects the distribution pattern finally learned by the network. In table 1, we have conducted experiments to determine the influence of sigma on the model results. 
\input{src/figandtable/abal_gaussiansigma.tex}

When the $\sigma$ is set to $0.3$ or $1$, the shape of distribution is either too narrow or too wide. When the distribution is too narrow, higher requirements are imposed on the model's predicted probability, which would lead to larger errors. If the $\sigma$ is too large, the targets becomes easier for the model to converge, but failed to further improve due to more values affect the final output.



\input{src/figandtable/abla_2.tex}
\subsubsection{Interpolation method}
To demonstrate the effectiveness of the proposed bilinear interpolation, we have conducted experiments to compare bilinear interpolation with trilinear interpolation. As shown in table \ref{Quantitative comparisons of different sigma}, bilinear interpolation has achieved better results with two methods, which aligns with our theory.

\subsubsection{Losses and Lambda $\lambda$}
We have also conducted experiments to compare the performance of different combination of losses and weight $\lambda$. As shown in table \ref{Quantitative comparisons of different sigma}, even though the cross-entropy(CE) loss has achieved only $0.94$, the network converges faster than trained with L1 loss. Regarding the combination of L1 and Cosine similarity(Cos), if the $\lambda$ is too large, the network would eventually collapse.

\subsubsection{Extended range}
At last, we conducted comparisons between with or without the extended range of disparity. As table \ref{Quantitative comparisons of different range} shown, it has a positive effect on the performance. 
\input{src/figandtable/abal_range.tex}





\subsection{Quantitative comparisons}
\input{src/figandtable/sceneflow_table.tex}
In this section, we compared with the SOTA methods and most relative methods on Sceneflow \cite{mayer_large_2016}, Kitti2012\citep{kitti2012} and Kitti2015\citep{kitti2015}.
In table \ref{tableSceneflow_q}, we compared with PDS\citep{PDS}, Acfnet\citep{AdaptiveUnimodal},PSMNet+\citep{chang_pyramid_2018}, GANet+LaC\citep{LocalSimilarity}, GANet+ADL\citep{Xu_2024_CVPR}. As shown, most methods utilize Top-k or other post-processing modules or integrate soft-argmax for supervised training. These methods employ additional modules, which leads to an increase in latency. In contrast, our method effectively improves the accuracy of the baseline and keeps the architecture unchanged, thus ensuring consistent and efficient inference.

\input{src/figandtable/kitti_table.tex}

The comparisons on Kitti are listed in table \ref{tablekitti}. As shown, our approach can effectively improve the results of all baselines. As the results indicate, the smaller the model, the greater the improvement. On MSN2D, we have achieved improvement of 0.54\%. And we also have obtained an improvement of 0.01\% on IGEV-stereo and achieved state-of-the-art results. Hence, it can be concluded that our method can effectively improve the generalization ability of the model, regardless of the model size. The improvement is particularly prominent on models with poor generalization.




\subsection{Qualitative comparisons}

\input{src/figandtable/sf_detail.tex}




Through experiments, we found that our \textit{Sampling-Gaussian} effectively improves the accuracy of the model to predicts small objects and details, as depicted in Figure \ref{figure_sf}.
The reason is that models trained with \textit{Soft-argmax} are prone to converge to the majority of the disparity, while details are relatively in the minority. On the other hand, our SG provides explicit supervision for all objects. Therefore, the model gains the ability to capture details.


\begin{figure}[h]
\includegraphics[width=0.99\linewidth]{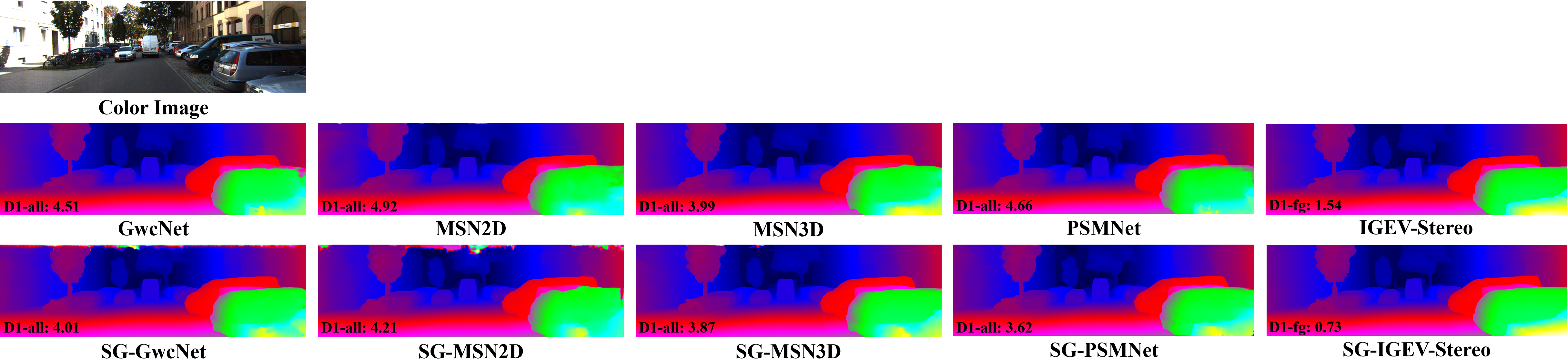}\\
\includegraphics[width=0.99\linewidth]{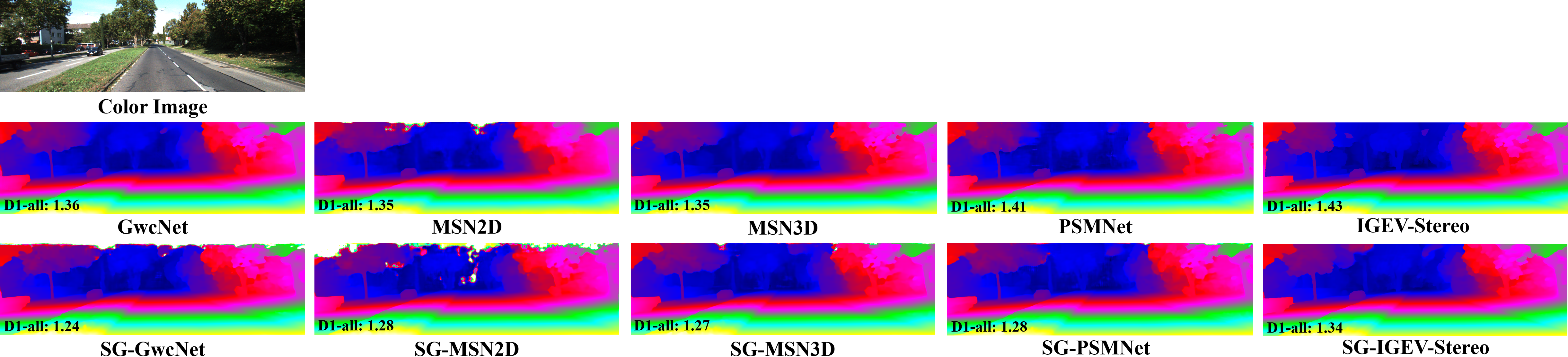}
\caption{Qualitative comparisons on Kitti2015}
\label{Qualitative comparisons on KITTI2015_2}
\end{figure}

In the first example in Fig. \ref{Qualitative comparisons on KITTI2015_2}, it is evident that all baselines trained with SG have gained the ability to capture details to different degrees. For instance, in the disparity of the right side van and the shape of the trees in the background. More of our results are available on the Kitti2012 and Kitti2015 leaderboard.

\subsection{Cross-domain generalization}
At last, we have conducted experiments to compare the cross-domain generalization ability of our methods. We have trained baselines on Sceneflow, and evaluate on Kitti2015 directly. Our method has improved the generalization ability of the baselines.
Qualitative results are available in appendix. 

\input{src/figandtable/cross.tex}

%% file: src/figandtable/abal_gaussiansigma.tex
\begin{table}[ht]
\centering
\caption{Quantitative comparisons on settings of $\sigma$ }
\begin{tabular}{ccccccc}
\toprule
$\sigma$   &$0.3$& $0.4$ & $\textbf{0.5}$ & $0.6$ & $0.7$ &  $1.0$ \\\midrule
PSMnet  & $2.526$  &$2.526$  &$ \textbf{0.625}$   & $0.631$  & $0.723$  &  $ 0.688$\\\bottomrule
\end{tabular}
\label{Quantitative comparisons of different}
\end{table}

%% file: src/figandtable/abla_2.tex
\begin{table}[!ht]
\centering
\caption{Quantitative comparisons on settings of $\sigma$.}
\begin{tabular}{l|cc|cccc}
\toprule
Base & Trilinear & Bilinear &Loss & $\lambda$   & EPE & D1 \\ \midrule
\multirow{2}{*}{MSN2D}&  \checkmark & ~ & L1 & / & 0.99 & 2.62 \\ 
~ & ~ &  \checkmark &  L1+Cos & 0.5 & \textbf{0.91} & \textbf{2.49} \\ \midrule
 \multirow{7}{*}{PSMNet} &\checkmark & & CE     & /   & 0.94 & 2.34 \\ 
 &  \checkmark & ~   & L1     & /   & 0.87 & 2.15 \\ 
 &  \checkmark & ~   & L1+Cos & 0.5 & 0.89 & 2.26 \\ 
 & ~ &  \checkmark  & L1+Cos &0.2 & 0.79 & 2.15 \\ 
 & ~ &  \checkmark  & L1+Cos & 1.0 & 1.23 & 2.86 \\ 
 & ~ &  \checkmark  & L1+Cos & 0.5 & \textbf{0.65} & \textbf{2.00} \\ 
\bottomrule
\end{tabular}
\label{Quantitative comparisons of different sigma}
\end{table}

%% file: src/figandtable/abal_range.tex
\begin{table}[ht]
\centering
\caption{Ablation study on disparity range}
\begin{tabular}{cccc}
\toprule
$d_{ext}$ &EPE& $<1$ & $<3$ \\ \midrule
0  & 0.69  & 6.72   & 2.32\\ 
16  & 0.65  & 5.31   & 2.00\\ 
\bottomrule
\end{tabular}
\label{Quantitative comparisons of different range}
\end{table}

%% file: src/figandtable/sceneflow_table.tex
\begin{table}[h]
\centering
\caption{Quantitative comparison on Sceneflow}
\begin{tabular}{lccccccc}
\toprule
Method  &EPE & D1 &  Params  & Supervision & Loss &Top-k &Time(s) \\
\midrule
PDS & 1.12 & 2.93 & 2.2 &Combined* & CE& Y &/\\
MSN2D & 1.14 & 2.83 & 2.23 &Soft-argmax &Smoothl1 & N& 0.10\\ 
PSMNet & 1.09 & 2.32 & 5.22 &Soft-argmax& Smoothl1& N& 0.41\\ 
PSMNet+& 1.02 & 3.12 & 2.32  &Laplacian & CE& Y & /\\
Acfnet& 0.87 & 4.31& /& Combined*& CE+Focal& N & 0.48\\
MSN3D & 0.80 & 2.10 & 1.77& Soft-argmax &Smoothl1 & N&0.53\\ 
GwcNet-g & 0.79 & 2.11 & 6.43 &Soft-argmax &Smoothl1 & N&0.32\\ 
GANet+LaC&0.72 &  6.52& 9.43&Combined* &L1+CE & Y& 1.72\\
GANet+ADL&0.50 & 1.81 & 9.43& Laplacian&L1+CE & Y& 1.72\\
IGEV-Stereo & 0.47 & 1.59 & 12.60 & Soft-argmax&L1 &N&0.37\\\midrule 
SG-MSN2D & 0.91 & 2.49 & 2.23 &Gaussian &L1+Cos &N&0.10\\ 
SG-PSMNet & 0.65 & 2.00 & 5.22&Gaussian &L1+Cos &N &0.41\\ 
SG-GwcNet-g & 0.71 & 2.09 & 6.43& Gaussian& L1+Cos&N &0.32\\ 
SG-MSN3D & 0.69 & 1.98 & 1.77&Gaussian &L1+Cos &N &0.53\\ 
SG-IGEV-Stereo & \textbf{0.47} & \textbf{1.58} & 12.60 &Gaussian&L1+Cos &N& 0.37\\
\bottomrule
\multicolumn{8}{l}{\small Combined*: combination of Soft-argmax and Laplacian}
\end{tabular}
\label{tableSceneflow_q}
\end{table}

%% file: src/figandtable/kitti_table.tex
\begin{table}[htbp]
\centering
\caption{The quantitative comparison on Kitti2012 and Kitti2015, the evaluation metrics are d1, $<2$ and $<3$ error rate($\%$). All are lower the better.}
\begin{tabular}{lllllllll}
\toprule
~ &  \multicolumn{3}{c}{Kitti2015-All} & \multicolumn{3}{c}{Kitti2015-Noc}  & \multicolumn{2}{l}{Kitti2012} \\ 
Method & $d1_{bg}$ & $d1_{fg}$ & $d1_{all}$ &  $d1_{bg}$ & $d1_{fg}$ & $d1_{all}$ & $<2$ & $<3$\\ \midrule
MSN2d\citep{shamsafar_mobilestereonet_2021} & 2.49 & 4.53 & 2.83 & 2.29 & 3.81 & 2.54 & $\backslash$ & $\backslash$ \\ 
PDSNet\cite{PDS} & 2.29 & 4.05 & 2.58 & 2.09 & 3.68 & 2.36 & 	4.65 & 2.53 \\
PSMnet\citep{chang_pyramid_2018} & 1.86 & 4.62 & 2.32 & 1.71 & 4.31 & 2.14 & 3.01 & 1.89 \\ 
PSMnet+CE\citep{Chen_2019_ICCV} & 1.54 & 4.33 & 2.14 & 1.70 &  3.90 &  1.93 & 2.81 & 1.81 \\
GwcNet-g\citep{guo_group-wise_2019} & 1.74 & 3.93 & 2.11 & 1.61 & 3.49 & 1.92 & $\backslash$ & $\backslash$ \\ 
MSN3d\citep{shamsafar_mobilestereonet_2021} & 1.75 & 3.87 & 2.10 & 1.61 & 3.50 & 1.92 & $\backslash$ & $\backslash$ \\ 
AAnet+\citep{xu_aanet_2020}& 1.65 & 3.96 & 2.03 & 1.49 & 3.66 & 1.85 & 2.96  & 2.04 \\ 
RAFT\citep{teed2021raft}& 1.48 & 3.46 & 1.81 & 1.34 & 3.11 & 1.63 & $\backslash$ & $\backslash$ \\ 
GANet\cite{GA_net} &  1.48 &  3.46  & 1.81 &  1.34 &  3.11 &  1.63 & 2.50 &  1.60 \\ 
ACVNet\citep{ACVnet} & 1.37 &  3.07 &  1.65 &  1.26 &  2.84 &  1.52  &  2.34 &  1.47 \\ 
RT-IGEV++ \citep{xu2024igevpp}& 1.48 & 3.37 & 1.79 & 1.34 & 3.17 & 1.64 & 2.51 & 1.68  \\ 
PSMNet+ADL\citep{Xu_2024_CVPR} & 1.44  & 3.25  & 1.74 &  1.30 &  3.04  & 1.59  &  2.17 &  1.42 \\ 
LEAstereo\cite{cheng_NAS_2020}&  1.40 &  2.91 &  1.65 &  1.29  & 2.65  & 1.51 &  2.39 &  1.45 \\ 
IGEV-stereo\citep{xu2023iterative} & 1.38 & 2.67 & 1.59 & 1.27 & 2.62 & 1.49 & 2.17  & 1.44 \\ \midrule
SG-MSN2d & 1.94 & 4.07 & 2.29 & 1.78 & 3.63 & 2.08 & 3.15 & 2.09 \\ 
SG-GwcNet-g & 1.73 & 3.88  &  2.09  &  1.59  &  3.55  &  1.92  &  2.89  &  1.95 \\ 
SG-PSMnet & 1.77 & 3.13 & 2.00 & 1.65 & 2.97 &1.87& 2.69 & 1.80 \\ 
SG-MSN3d & 1.61 &  3.81  &  1.98  &  1.48  &  3.55  &  1.82  &  2.62  &  1.74 \\ 
SG-IGEV-stereo & 1.40  & \textbf{2.50} & \textbf{1.58} &  1.30  & \textbf{2.48} & 1.50  & \textbf{2.12} & \textbf{1.39} 
\\\bottomrule
\end{tabular}
\label{tablekitti}
\end{table}

%% file: src/figandtable/sf_detail.tex
\begin{figure}[ht]
\centering
\includegraphics[width=1\linewidth]{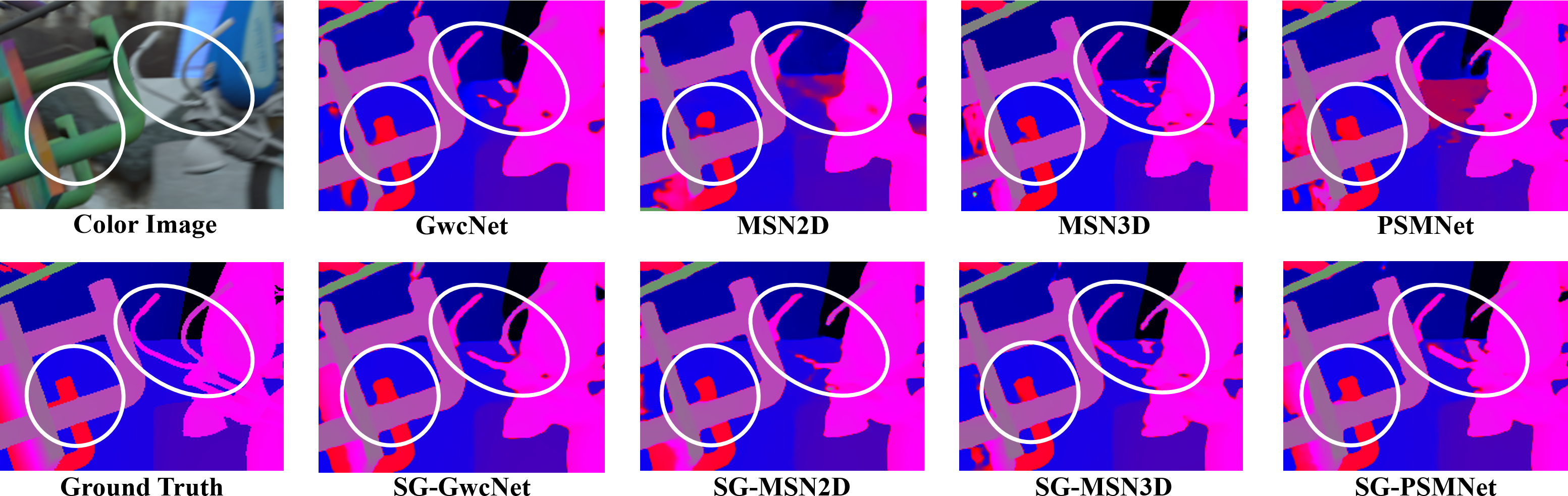}
\caption{Qualitative comparisons on Sceneflow}
\label{figure_sf}
\end{figure}

%% file: src/figandtable/cross.tex
\begin{table}[h]
\centering
\caption{Cross-domain generalization evaluation}
\begin{tabular}{lcccc|lcccc}
\toprule
~ &\multicolumn{4}{c|}{Kitti2015-ALL}  & &\multicolumn{4}{c}{Kitti2015-ALL}   \\ 
Base   & EPE  & $>1$ & $>2$ &$ >3$& Ours   & EPE  & $>1$ & $>2$ &$ >3$ \\ \midrule
MSN2D    & 5.03 & 56.1   & 33.7 &  24.4 &SG-MSN2D & \textbf{1.53} &\textbf{48.2}   & \textbf{22.2} &  \textbf{12.5} \\ 
MSN3D    & 29.4 & 72.2   & 57.7 &  50.0 &SG-MSN3D & 22.5 & 53.7   & 26.3 &  17.3\\ 
PSMNet   & 21.1 & 88.6   & 64.7 &  48.8 & SG-PSMNet  & 24.6 & 78.0   & 65.0 &  57.2\\ 
\bottomrule
\end{tabular}
\end{table}

%% file: src/chap6_conculsion.tex
\section{Conclusions}


In this paper, we introduce a novel training method \textit{Sampling-Gaussian} for stereo matching. We have solved the fundamental problems of previous distribution-based method by extend the disparity range and bilinear interpolation. Moreover, we interpret the learning process as minimizing the distance in the vector space, and proposed a combined loss.
Through comprehensive comparisons with five baseline methods, we demonstrate that our \textit{Sampling-Gaussian} achieves improvements through all the methods, and fulfill our goal in proposing an effective and easy to implement method.
In the future, we are going to study the generalization ability of stereo matching networks in order to solve the applications in real-life.


%% file: src/chapF_appendix.tex
\section{Appendix}



\subsection{Full equation of Equ. \ref{softargmaxequ}}
The first part is the full equation of Equ \ref{softargmaxequ}.
\begin{equation}
\begin{aligned}
\frac{\partial L}{\partial e^{z_{i}}}&=\frac{
\partial L}{\partial d}\frac{\partial d}{\partial e^{z_{i}}}
\\&=
\frac{\partial L}{\partial d}(i\frac{e^{z_{i}}}{\sum e^{*}}(1-\frac{e^{z_{i}}}{\sum e^{*}})+\sum_{j\neq i}j(-\frac{ e^{z_{j}}}{\sum e^{*}}*\frac{e^{z_{i}}}{\sum e^{*}}))
\\&=
\frac{\partial L}{\partial d}(i\frac{e^{z_{i}}}{\sum e^{*}}+i(-\frac{e^{z_{i}}}{\sum e^{*}}*\frac{e^{z_{i}}}{\sum e^{*}})+\sum_{j\neq i}j(-\frac{ e^{z_{j}}}{\sum e^{*}}*\frac{e^{z_{i}}}{\sum e^{*}}))
\\&=
\frac{\partial L}{\partial d}(i\frac{e^{z_{i}}}{\sum e^{*}}+\sum_{j}(-\frac{ e^{z_{j}}}{\sum e^{*}}*\frac{e^{z_{i}}}{\sum e^{*}}))
\\&=
\frac{\partial L}{\partial d}(\frac{e^{z_{i}}}{\sum e^{*}}(i-\underline{\sum j*\frac{ e^{z_{j}}}{\sum e^{*}}}))
\\&=
\frac{\partial L}{\partial d}(\frac{ e^{z_{i}}}{\sum e^{*}}(i-d))
 \end{aligned}
 \end{equation}
the part with underline is the equation of soft-argmax Equ. \ref{softargmax},

\subsection{Python Implementation}
This is the python implementation of \textit{Sampling-Gaussian}. 
\label{Python Implementation}
\begin{figure}[ht]
\centering
\includegraphics[width=0.9\linewidth]{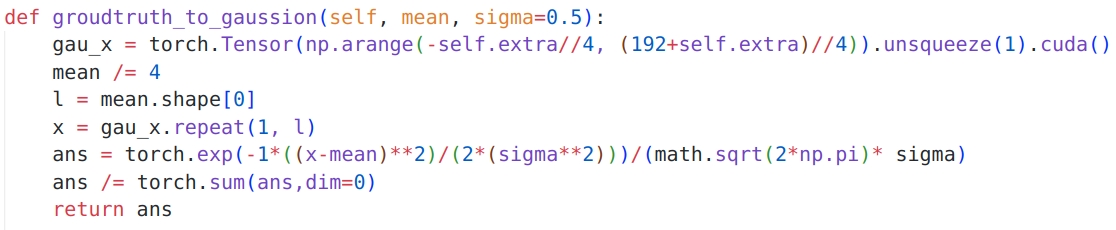}
\end{figure}

\subsection{Probabilities of Sampling-Gaussian}
\begin{table}[ht]
    \centering
    \caption{The accuracy of the Sampling-Gaussian's cumulative possibility and expectation.
    }
    \begin{tabular}{c|c|c}
    \toprule
    $\mu$ & $1-\sum_{x} p$    & $\mu - \sum_{x} d*p$\\\midrule
    4 &$ 0.005296 $&$ -0.37134$ \\
    5 &$ 0.004317 $&$ -0.02964$ \\
    6 &$ 3.14e-05 $&$ -0.00178$ \\
    7 &$ 1.10e-06 $&$ -6.2e-05$ \\
    8 &$ 2.37e-08 $&$ -1.3e-06$ \\
    7 &$ 1.10e-06 $&$ -6.2e-05$ \\
    8 &$ 2.37e-08 $&$ -1.3e-06$ \\
    9 &$ 3.07e-10 $&$ -1.8e-08$ \\
    10 &$ 2.39e-12 $&$ -1.4e-10$ \\
    11 &$ 1.09e-14 $&$ -6.8e-13$ \\
    12 &$ 0.00 $&$ 7.10e-15$ \\
    15 &$ 0.00 $&$ 0.00.0$ \\
    20 &$ 0.00 $&$ 7.10e-15$ \\
    \bottomrule
    \end{tabular}
    \label{Sampling-Gaussian's cumulative possibility}
    \end{table}
Let's review the equation \ref{sampling-gaussian}.
First, the probability density function of the discretized Gaussian distribution is defined as 
\begin{equation}
    q(x)=\frac{1}{\sigma*\sqrt{2\pi}}e^{-\frac{(x-\mu)^2}{2\sigma^{2}}}
    \label{discretergaussian}
\end{equation}

The Riemann sum  of the equation \ref{discretergaussian} is
\begin{equation}
\int_{a}^{b}e^{-\frac{(x-\mu)^{2}}{2\sigma ^{2}}} dx \approx \frac{1}{2}(f(x_{0})+2f(x_{1})\dots +2f(x_{N-1})+f(x_{N}))
\label{Riemann sum}
\end{equation}
We further evaluate the summation of probability of Equ. \ref{Riemann sum}.
Thus, we need to evaluate the \textit{Sampling-Gaussian}'s cumulative possibility.
As shown in Table \ref{Sampling-Gaussian's cumulative possibility}. The table shows, that the cumulative possibility is not strictly equals to $1$. However, the probabilities predicted by the network is strictly equals to $1$ due to the softmax operation.
Therefore, in Equ. \ref{sampling-gaussian}, the probabilities is divided by the summation of the probabilities. Thus, the summation is strictly equals to $1$.

The table \ref{Sampling-Gaussian's cumulative possibility} shown the range inside the $[0,d_{max})$. Which illustrate the reason of why $d_{ext}$ is needed. Moreover, as depicted in table \ref{Sampling-Gaussian's cumulative possibility}.  The cumulative possibility is not always equals to 1. Therefore, the division by the summation of the probabilities is an effective to strictly restrict the probability equals to $1$.





\begin{figure}[ht]
    \centering
    \includegraphics[width=0.40\linewidth]{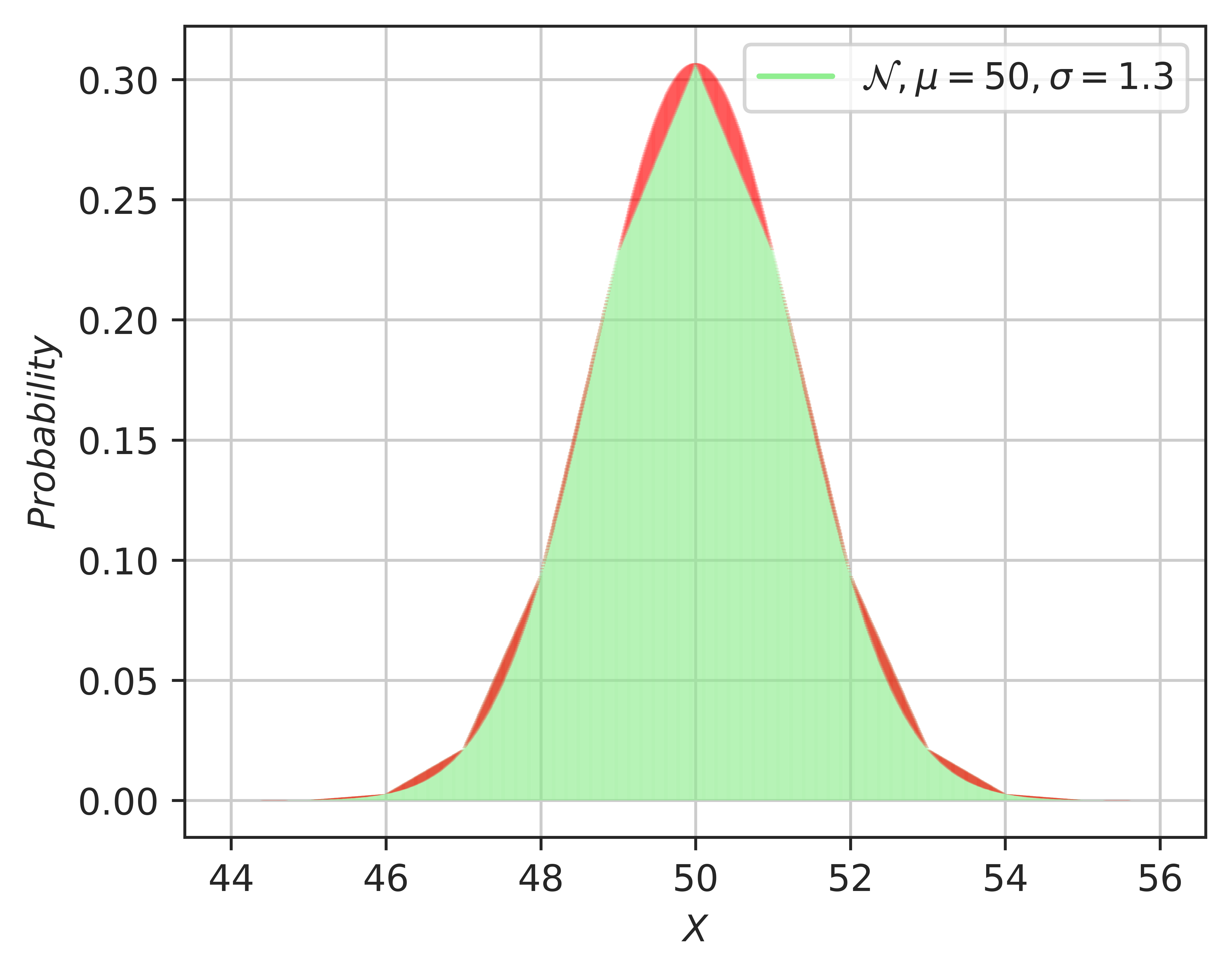}
    \caption{The green region represents the integral of Equ. \ref{discretergaussian}, while the red area denotes the difference between the integrals and cumulative probability of $SG$.}
\end{figure}

\subsection{More analysis and properties}
During the research, we have discovered that our \textit{Sampling-Gaussian} possesses two interesting properties: Firstly, within a certain range of $\sigma\in [0.9,1.7]$, its sum approximates to $1$. Secondly, its expectation is equal to $\mu$.
The first property: that a finite integration of Gaussian distribution is defined by $\int_{a}^{a+1}e^{-\frac{(x-\mu)^{2}}{2\sigma ^{2}}} dx$.
The numerical integration is
\begin{equation}
\int_{a}^{a+1}e^{-\frac{(x-\mu)^{2}}{2\sigma ^{2}}} dx \approx \frac{1}{2}(e^{-\frac{(a-\mu)^{2}}{2\sigma ^{2}}}+e^{-\frac{(a+1-\mu)^{2}}{2\sigma ^{2}}}).
\end{equation}
Let $\{x_{k}\}$ be a partition of $[a,b]$, $a=x_{0}<x_{1}\dots<x_{N-1} <x_{N}=b$, and the partition has a regular spacing $x_{k}-x_{k-1} = 1$. The approximation formula can be simplified as
$\int_{a}^{b}e^{-\frac{(x-\mu)^{2}}{2\sigma ^{2}}} dx \approx \frac{1}{2}(f(x_{0})+2f(x_{1})\dots +2f(x_{n-1})+f(x_{n}))$. Let $a= -\infty$, $b= \infty$, then we have
\begin{equation}
\frac{1}{\sigma \sqrt{2\pi}}\int_{-\infty}^{\infty}e^{-\frac{(x-\mu)^{2}}{2\sigma ^{2}}} dx\approx \frac{1}{\sigma \sqrt{2\pi}}\sum_{x\in \mathbb{Z}}e^{-\frac{(x-\mu)^{2}}{2\sigma ^{2}}}.
\label{lemma42equ}
\end{equation}




Second property: For simplicity, let $f(x) = e^{-\frac{(x-\mu)^{2}}{2\sigma ^{2}}}$.
$\forall x>\mu, \partial f/\partial x<0$.
Let $0\leqslant t \leqslant 1, i<j$, 
$\forall x\in \{x_{i}|x\geqslant b, x_{i}\in \mathbb{Z} \}$, $f(x)$ satisfies $f(x_{i}+t*(x_{j}-x_{i}))\leqslant f(x_{i})+t[f(x_{j})-f(x_{i})]$. Therefore, the numerical integration $\frac{1}{2}(x_{n}-x_{1})\cdot (f(x_{i})+f(x_{n}))=\epsilon$ satisfies $\epsilon>\sum_{x=b}^{\infty}f(x)>0$.
Based on our numerical analysis, when $\delta=5$, $\epsilon<10^{-5}$, the 
\begin{equation}
    \frac{1}{\sigma \sqrt{2\pi}}\sum_{x\in \mathbb{Z}}f(x) - 2\epsilon = \frac{1}{\sigma \sqrt{2\pi}}\sum_{x=\mu-b}^{\mu+b}f(x) \approx 1.
\end{equation}
Let $\mu \in (0,d_{max}), \sigma \in [0.5,1.0]$, the expectation
\begin{equation}
E(x|\mu) = \sum_{x=0}^{d_{max}}\frac{1}{\sigma \sqrt{2\pi}}e^{-\frac{(x-\mu)^{2}}{2\sigma ^{2}}} \approx \mu.
\end{equation}
let $\mu \in (5,d_{max}-5), x^{*}\in \{x^{*}<0 \cup x^{*}\geqslant d_{max}\}$. Then $E(x^{*}|\mu)\approx 0$.
Given the finite range of disparity $[0,d_{max})$, by subtracting the $E(x^{*}|\mu)$ from the $E(x)$.
We have also conducted experiments to quantize the error of the expectations and the error ranges from $10^{-5}$ to $10^{-12}$.









\subsection{Training and inference}

The training and inference process is illustrated as:

\begin{algorithm}[h]
\label{alg1}
\caption{Training with \textbf{sampling-Gaussian}}
\begin{algorithmic}[1]
\renewcommand{\algorithmicrequire}{\textbf{Input:}}
\renewcommand{\algorithmicensure}{\textbf{Output:}}
\REQUIRE left, right image $I_{l}, I_{r}$, ground truth $\hat{d}$, sampling-Gaussian $f$, threshold $T$, set $S_{x}$.
\ENSURE Network $N$.
\WHILE{$loss>T$}
\STATE $y \gets N(I_{l}, I_{r})$
\STATE $d \gets Softmax(y)$  
\STATE $\hat{d} \gets f(x=S_{x}|\mu=\hat{d})$
\STATE $loss \gets L1(d, \hat{d})-0.5*cos(d, \hat{d})$
\STATE update network by backpropagation
\ENDWHILE
\end{algorithmic}
\end{algorithm}




\subsection{More quantitative comparisons}
\input{src/figandtable/qualitative_comp2.tex}
\input{src/figandtable/qualitative_comp1.tex}
\input{src/figandtable/app_fig.tex}


\subsection{The results on Kitti2012 and Kitti2015}
At last, we provide the URL of our submitted results on Kitti leaderboard.
\href{https://www.cvlibs.net/datasets/kitti/eval_scene_flow_detail.php?benchmark=stereo&result=23c583cd9839dd0f12ba3bebaa30771c340684d8}{SG-PSMNet on Kitti2015},
\href{https://www.cvlibs.net/datasets/kitti/eval_scene_flow_detail.php?benchmark=stereo&result=e8c33b9f5a4eb3450c383905ea009ac8e0df31ba}{SG-MSN2D on Kitti2015},
\href{https://www.cvlibs.net/datasets/kitti/eval_scene_flow_detail.php?benchmark=stereo&result=03ec9abd0e870587cb0785f2d11e047a8bd010a4}{SG-MSN3D on Kitti2015},
\href{https://www.cvlibs.net/datasets/kitti/eval_scene_flow_detail.php?benchmark=stereo&result=6df61ea412715e6ee615e2e451bff8e384b61326}{SG-GwcNet-g on Kitti2015},
\href{https://www.cvlibs.net/datasets/kitti/eval_scene_flow_detail.php?benchmark=stereo&result=b759b993d8cdce000d8edd7360913160cf8a6935}{SG-IGEV on Kitti2015}.
\href{https://www.cvlibs.net/datasets/kitti/eval_stereo_flow_detail.php?benchmark=stereo&error=3&eval=all&result=ce9458fc723a07dc1d0ca6872e95923fad5f9aa2}{SG-PSMNet on Kitti2012},
\href{https://www.cvlibs.net/datasets/kitti/eval_stereo_flow_detail.php?benchmark=stereo&error=3&eval=all&result=19e8e20452881f5b71056e2b5aa6667b4c75c2f2}{SG-MSN2D on Kitti2012},
\href{https://www.cvlibs.net/datasets/kitti/eval_stereo_flow_detail.php?benchmark=stereo&error=3&eval=all&result=49679a12953317297ec44772e2f8a9cda0ab60e2}{SG-MSN3D on Kitti2012},
\href{https://www.cvlibs.net/datasets/kitti/eval_stereo_flow_detail.php?benchmark=stereo&error=3&eval=all&result=6c51e3a89e60413c6c7b74d13d611db2f173b0d5}{SG-IGEV on Kitti2012}.

%% file: src/figandtable/qualitative_comp2.tex
\begin{figure}[h]
    \centering
    \begin{minipage}[t]{0.19\textwidth}
    \centering
    \includegraphics[width=0.97\textwidth]{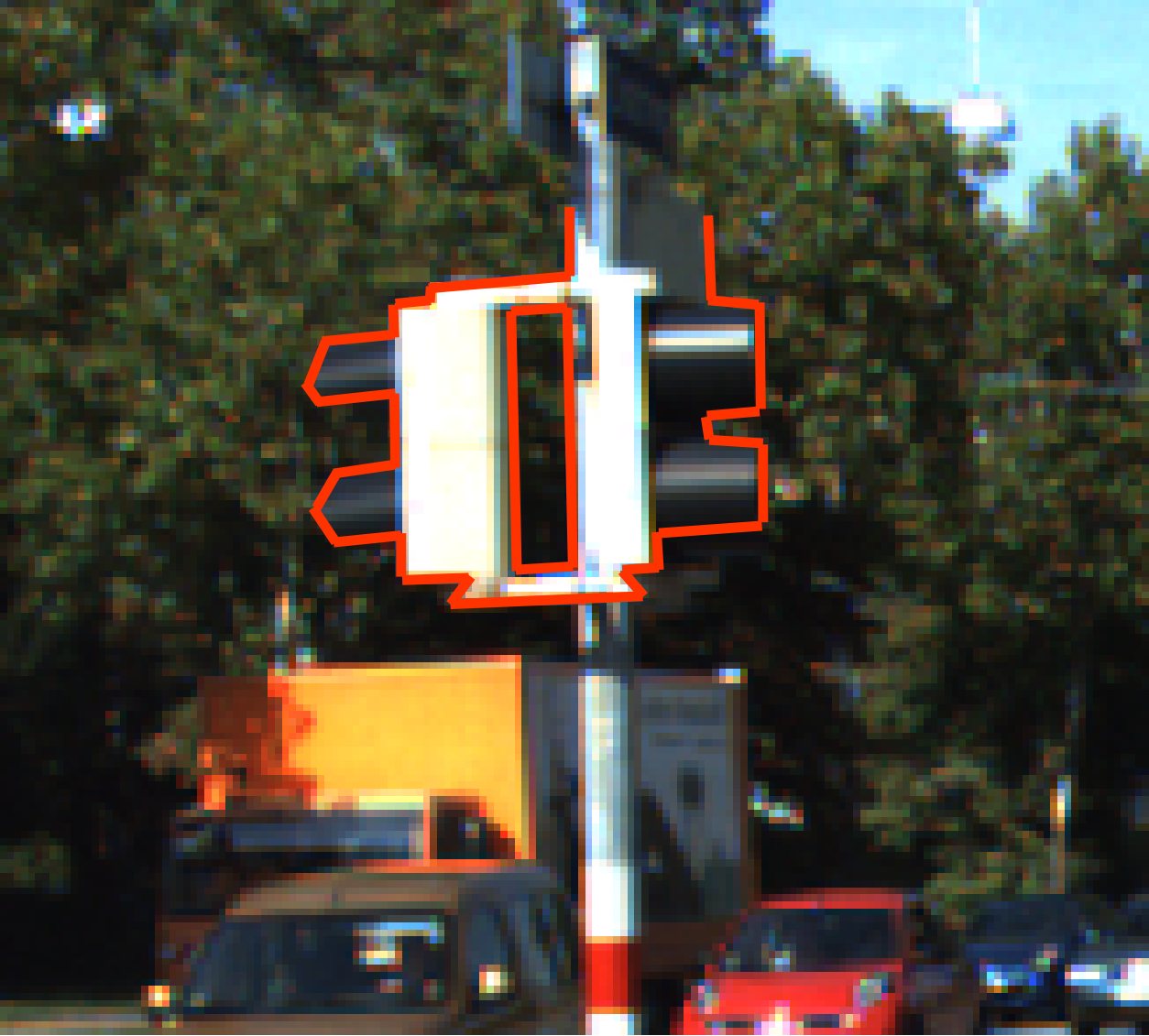}\\
    \includegraphics[width=0.97\textwidth]{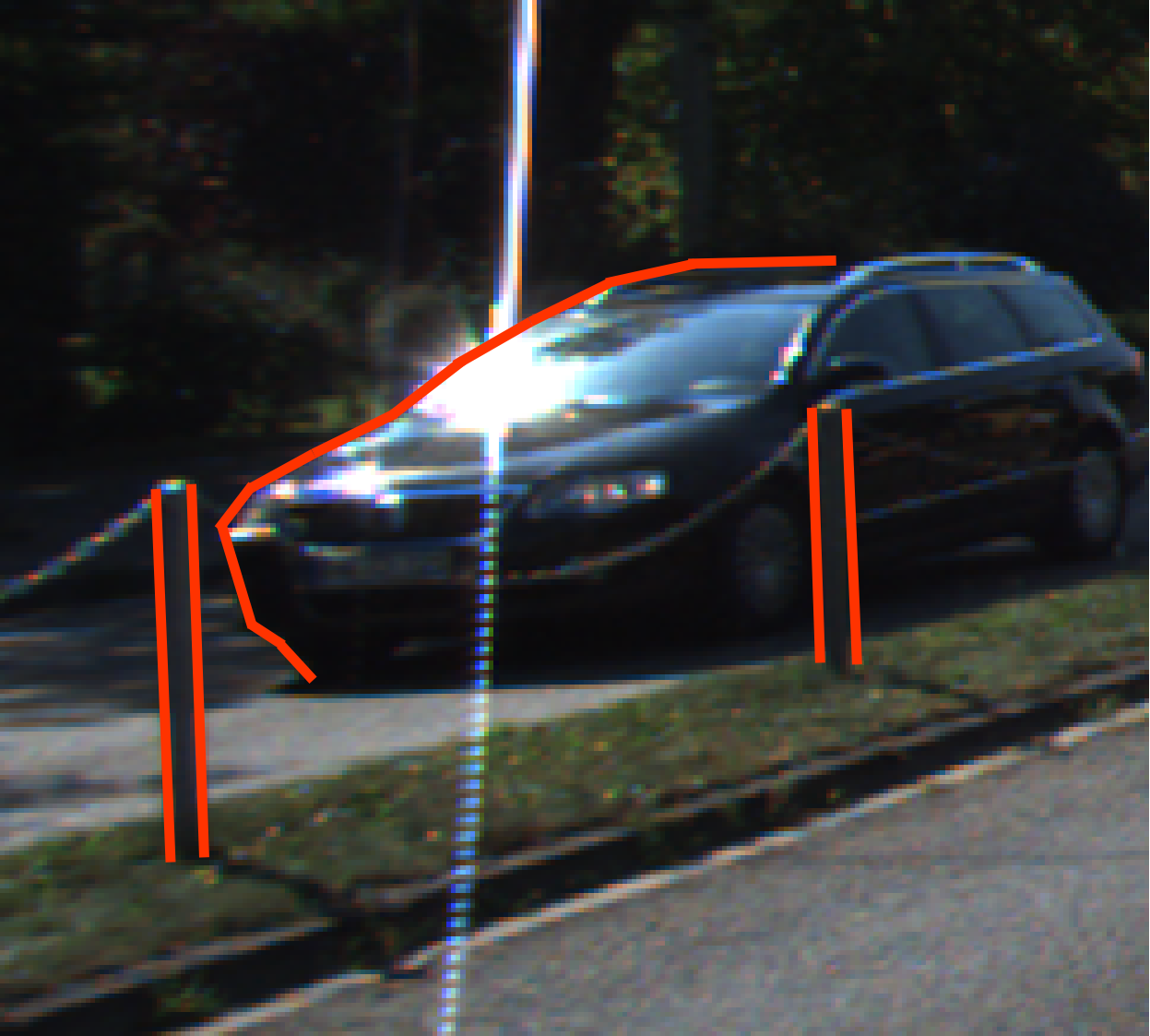}
    Color Image
    \end{minipage}
    \begin{minipage}[t]{0.19\textwidth}
    \centering
    \includegraphics[width=0.97\textwidth]{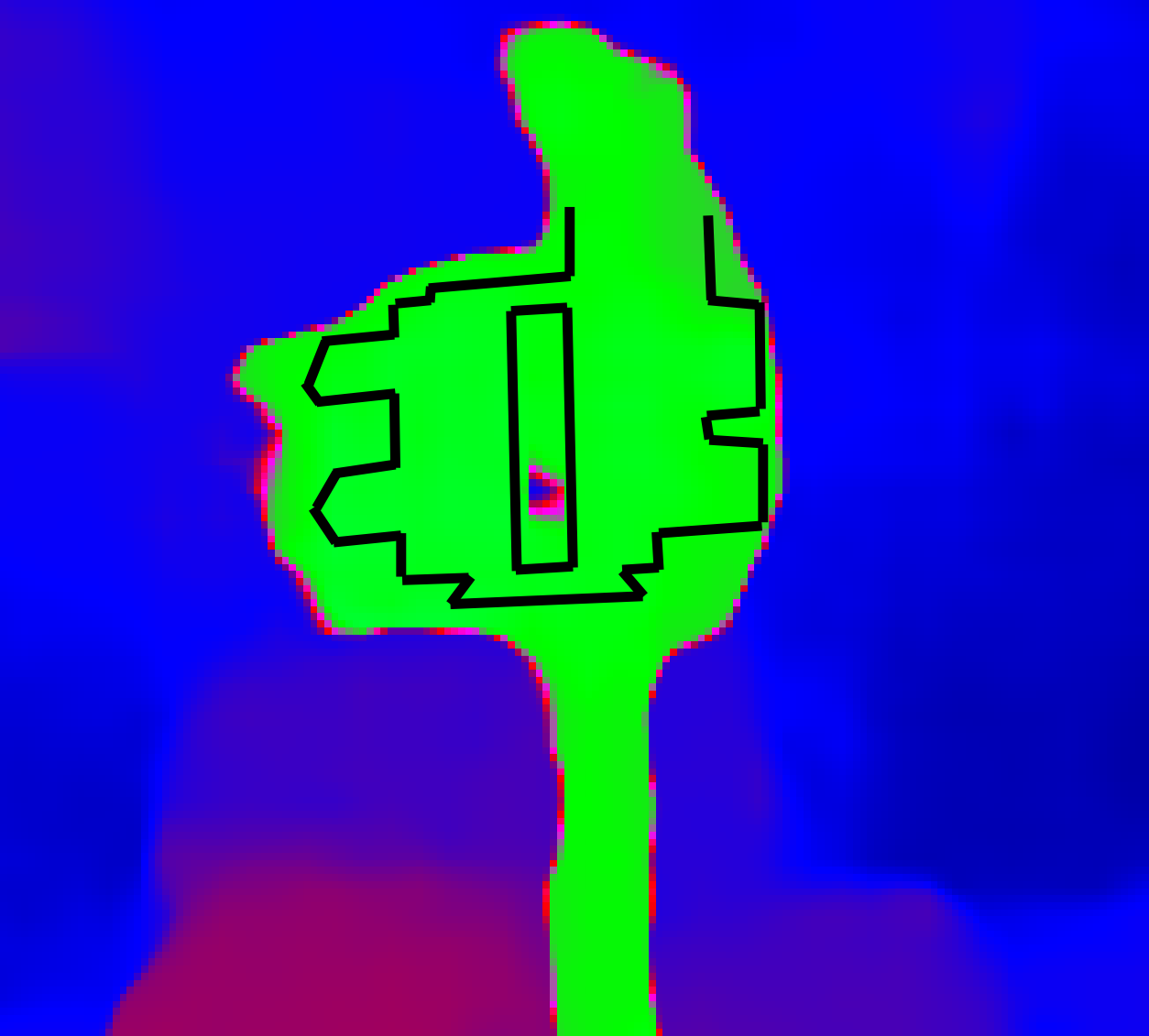}\\
    \includegraphics[width=0.97\textwidth]{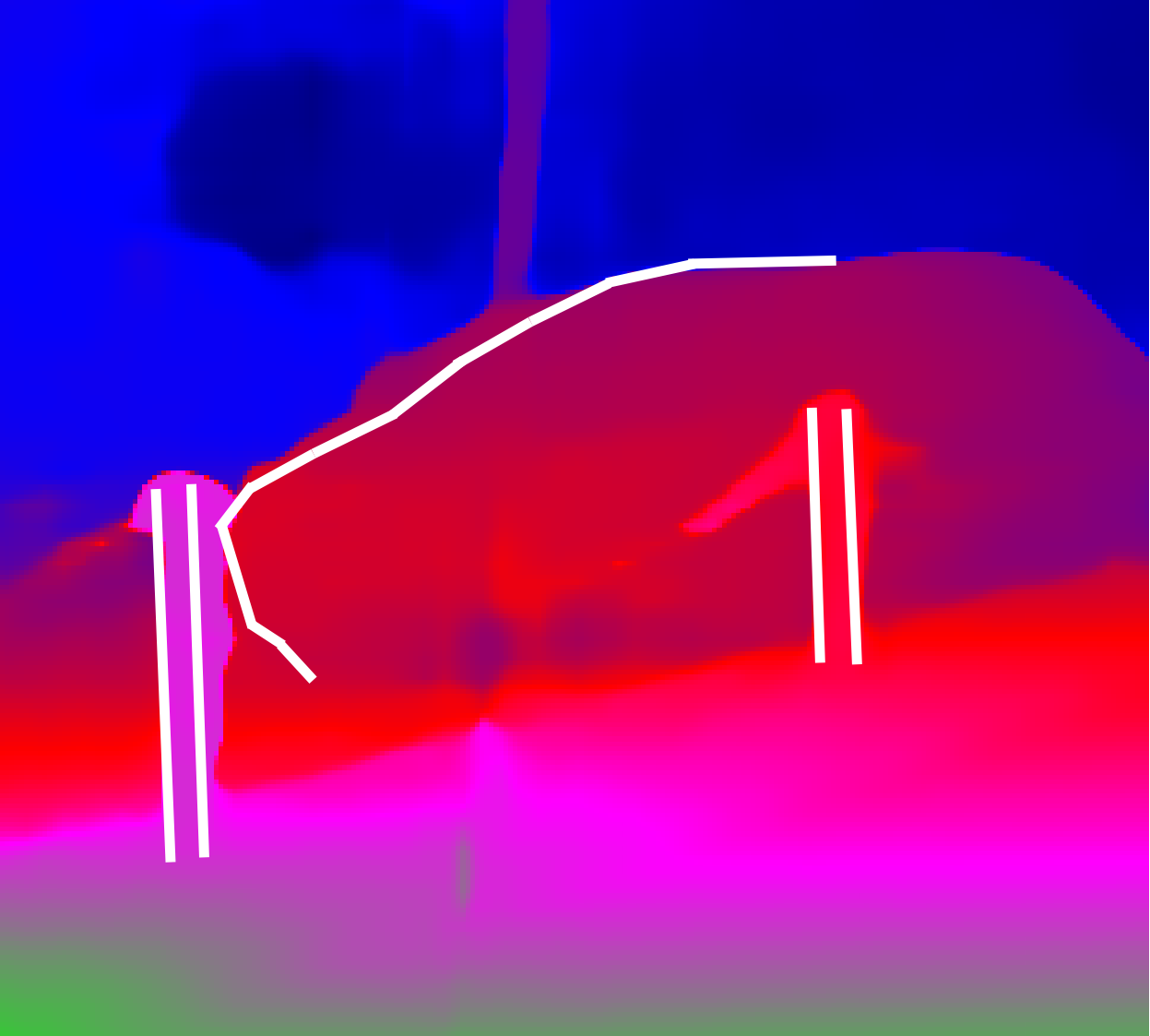}
    PSMNet
    \end{minipage}
    \begin{minipage}[t]{0.19\textwidth}
    \centering
    \includegraphics[width=0.97\textwidth]{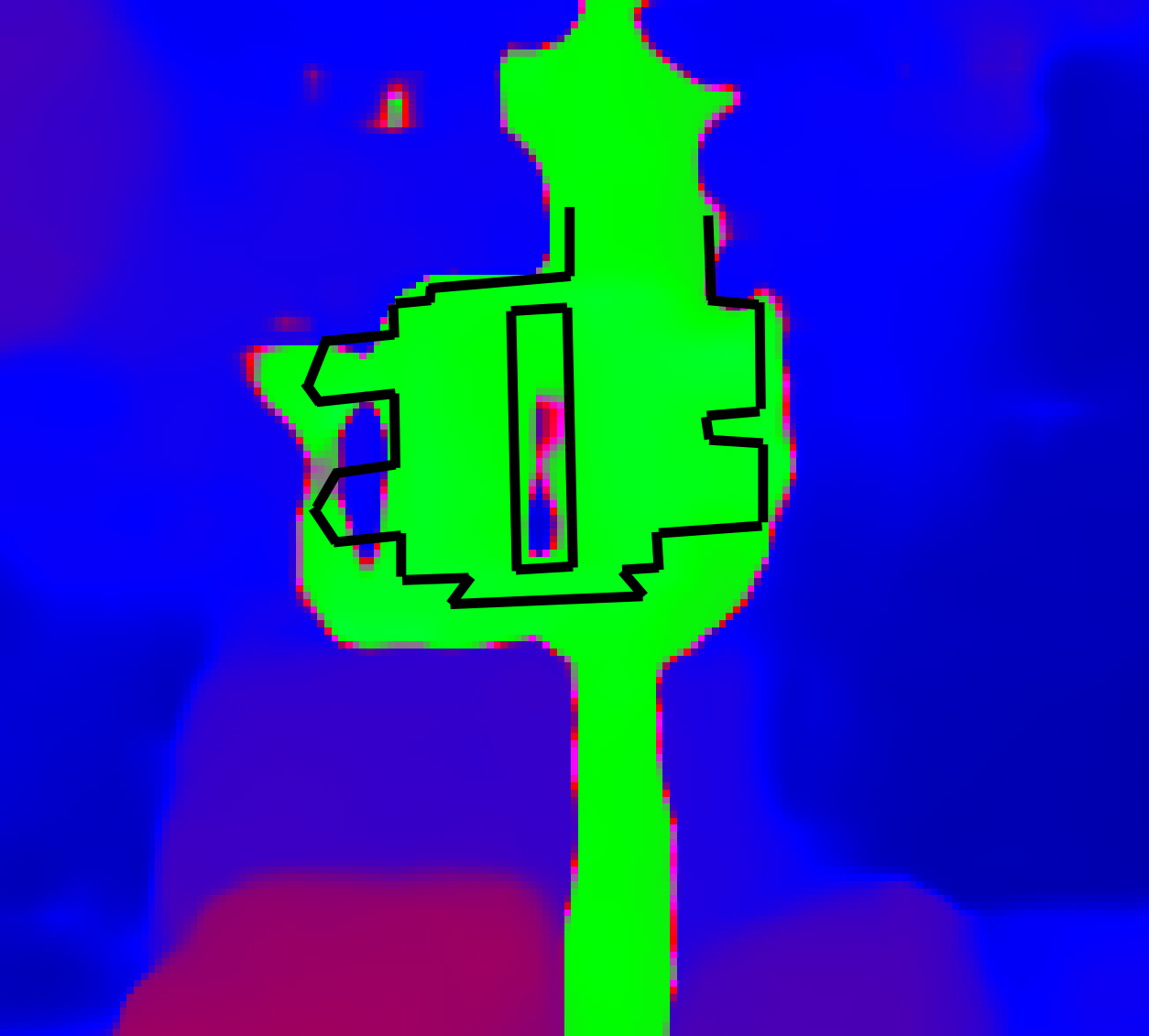}\\
    \includegraphics[width=0.97\textwidth]{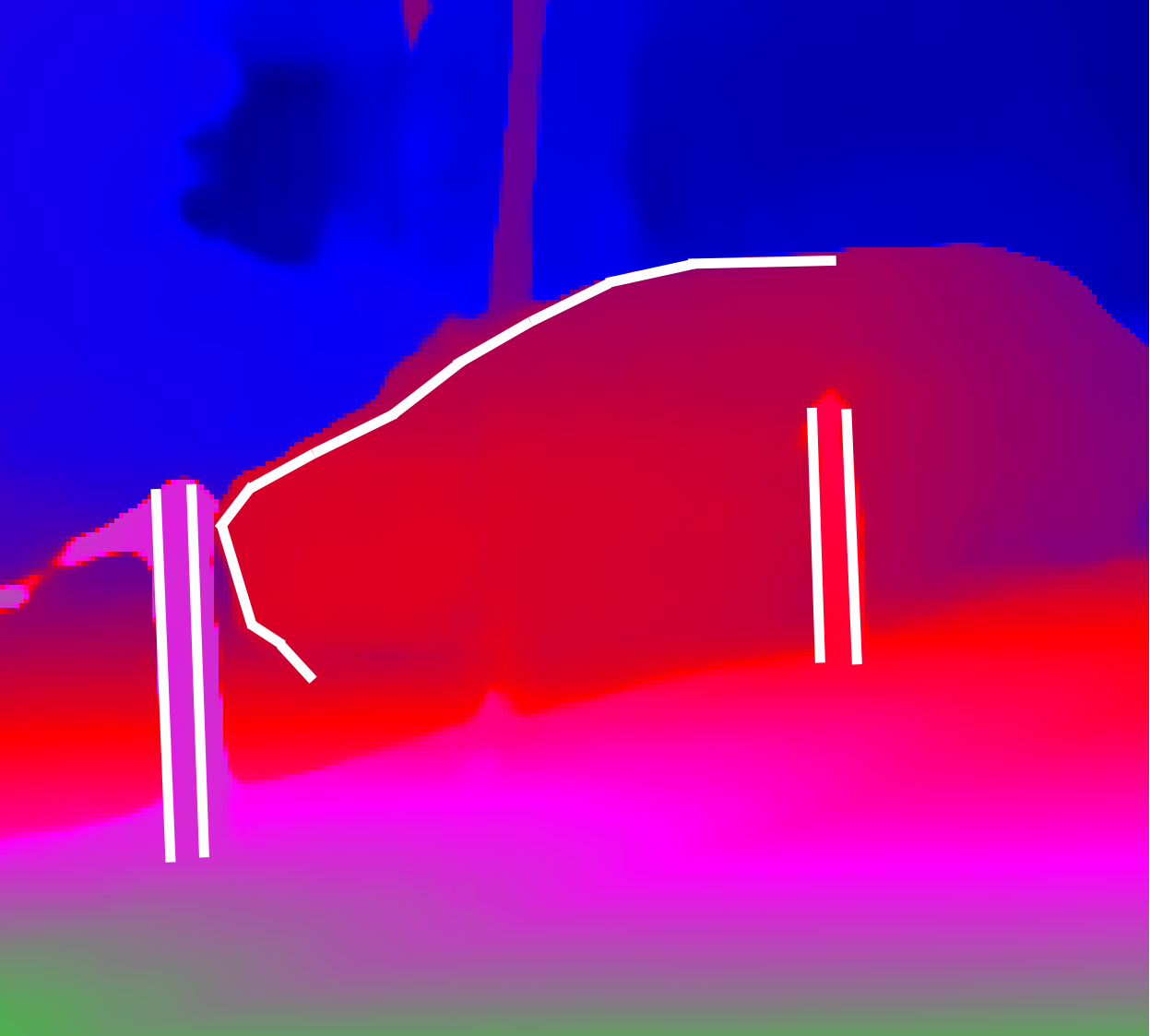}\\
    SG-PSMNet
    \end{minipage}
    \begin{minipage}[t]{0.19\textwidth}
    \centering
    \includegraphics[width=0.97\textwidth]{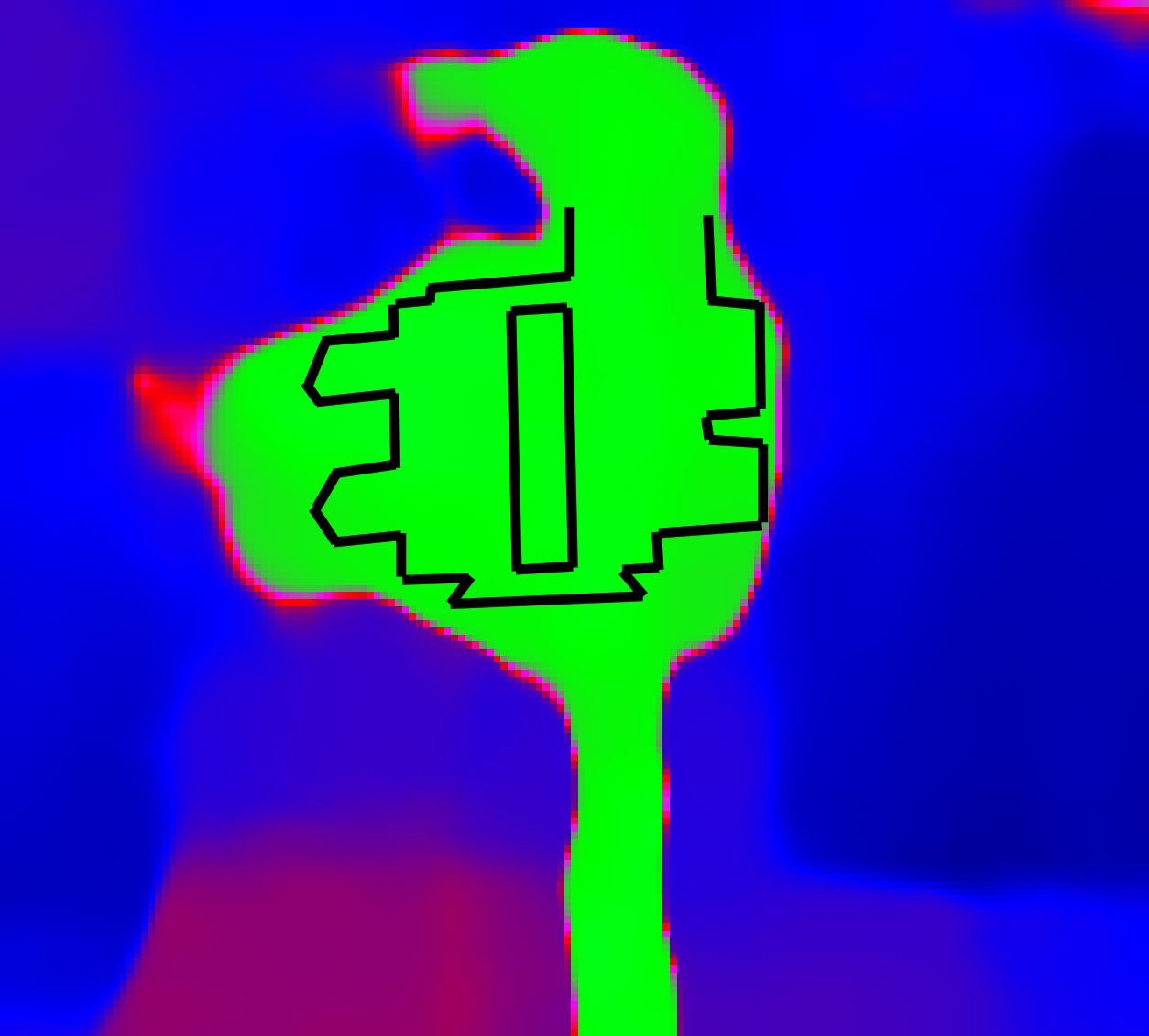}\\
    \includegraphics[width=0.97\textwidth]{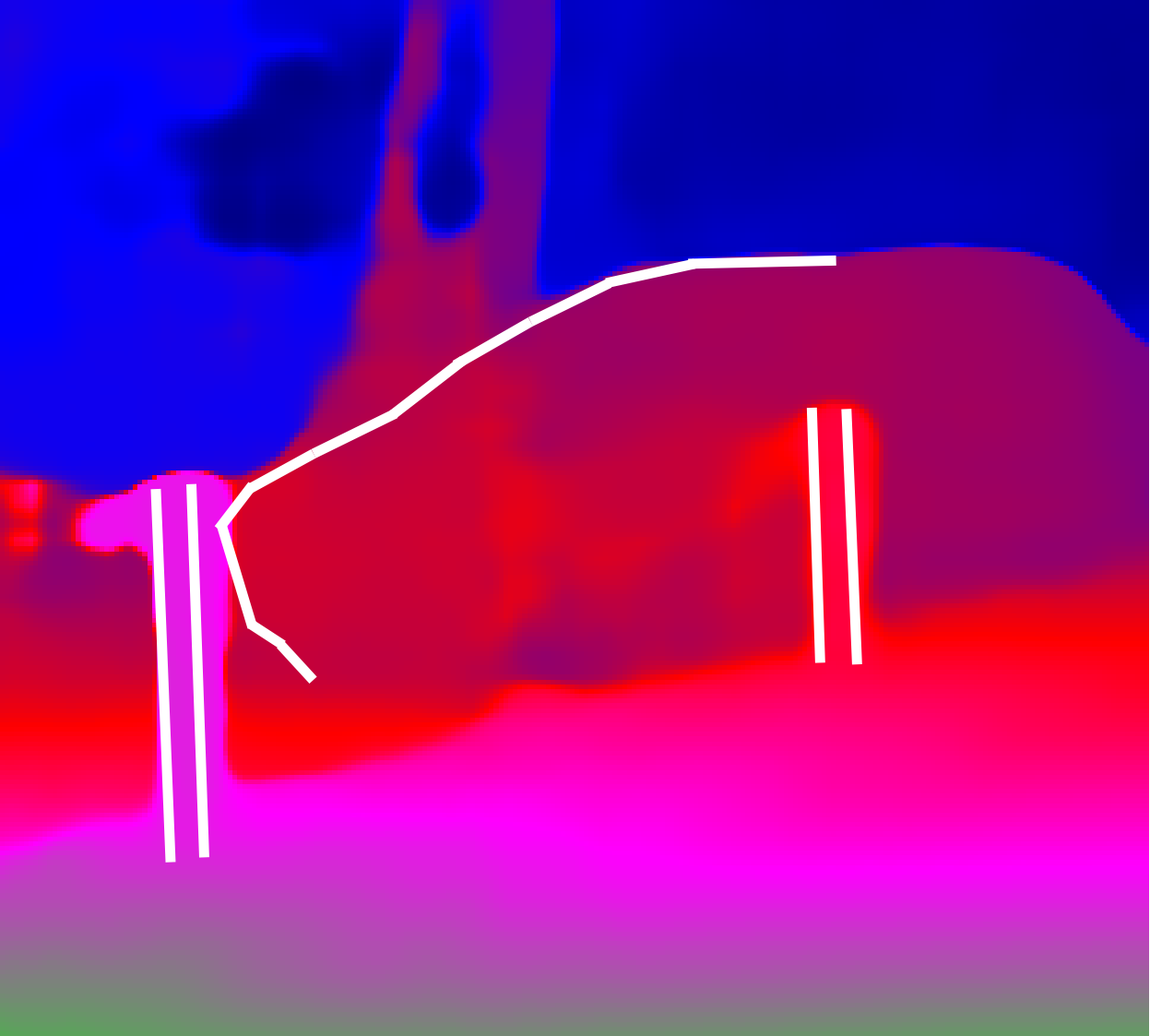}
    MSN2D
    \end{minipage}
    \begin{minipage}[t]{0.19\textwidth}
    \centering
    \includegraphics[width=0.97\textwidth]{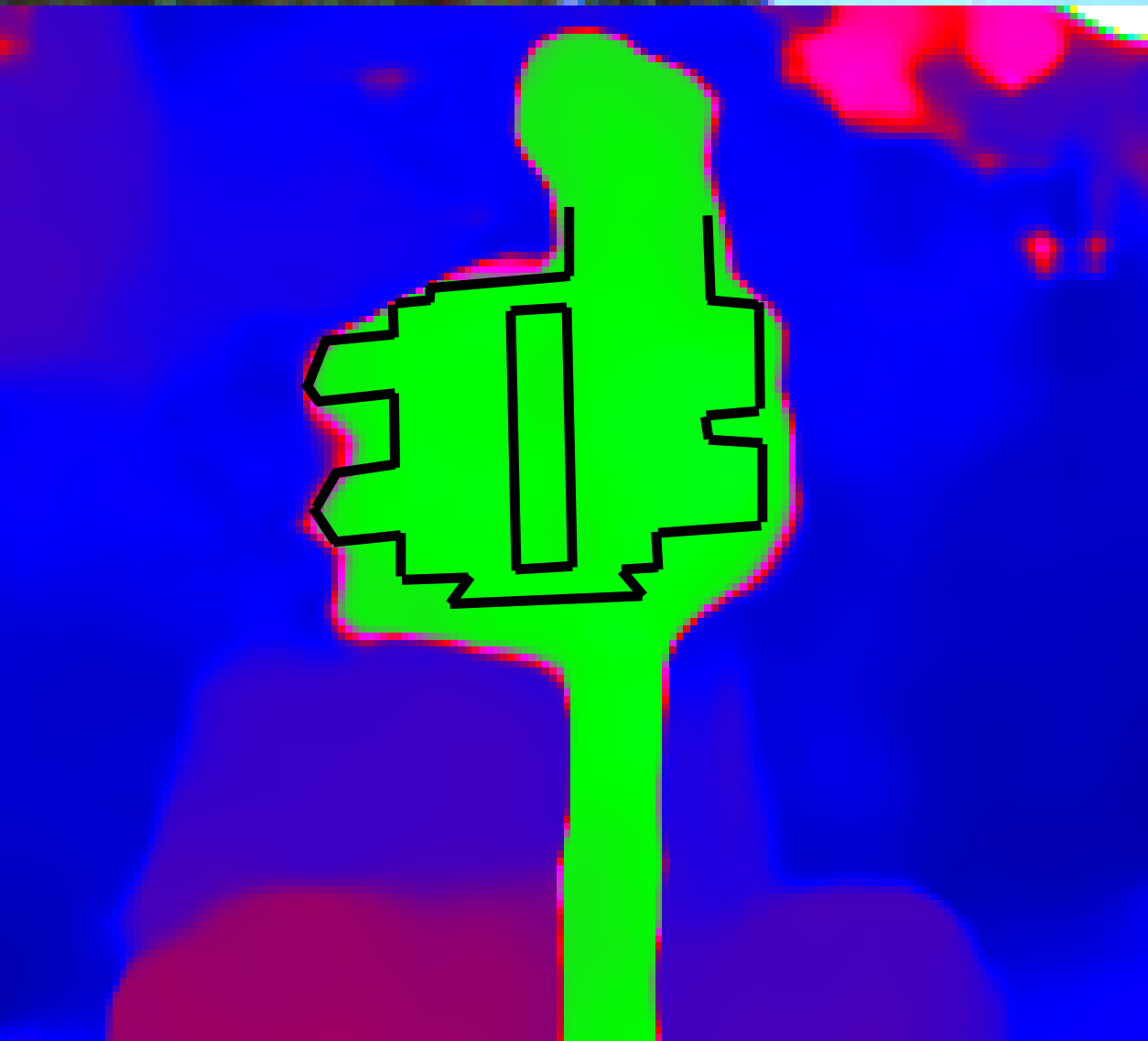}\\
    \includegraphics[width=0.97\textwidth]{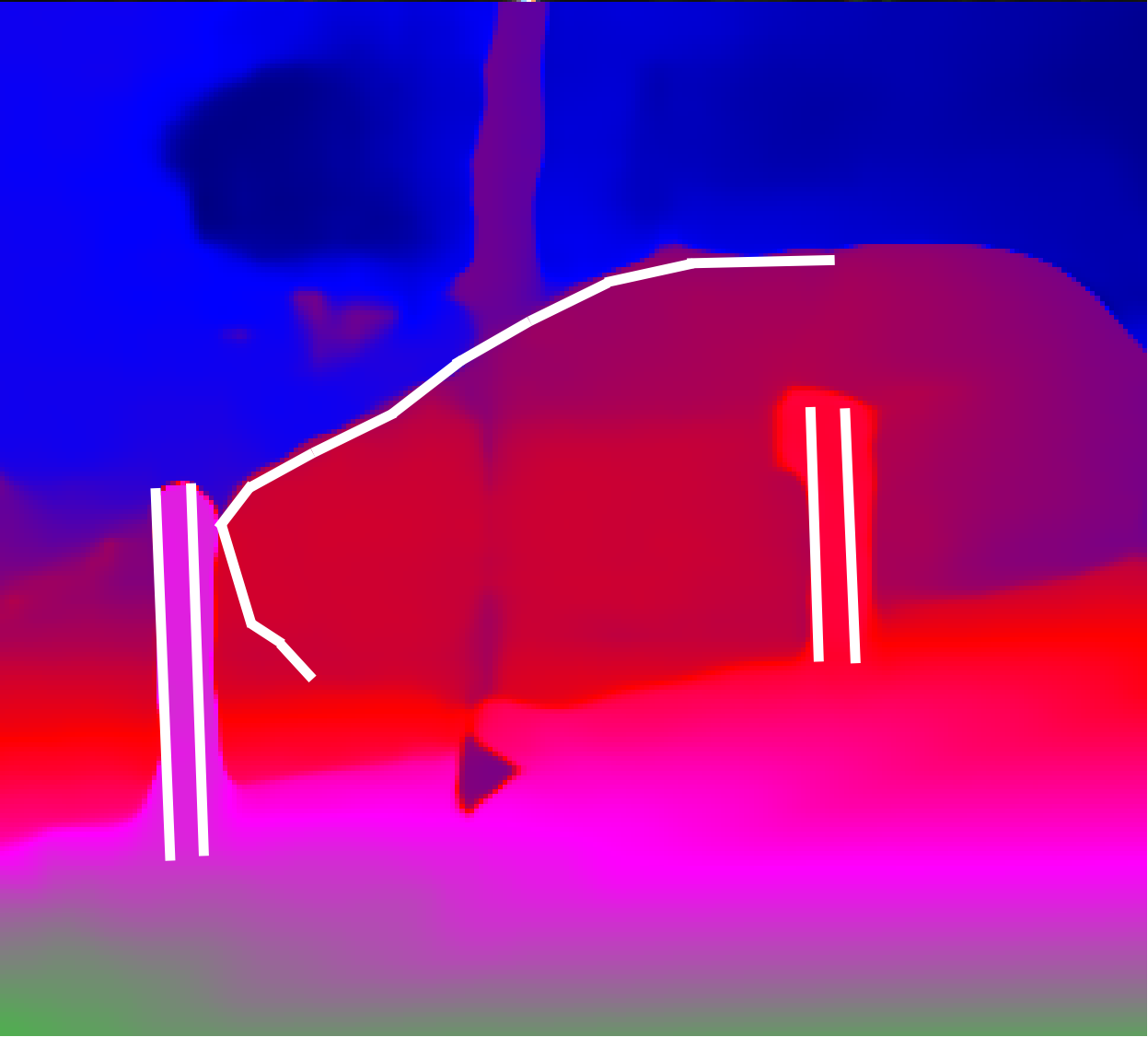}
    SG-MSN2D
    \end{minipage}
    \caption{Qualitative comparisons on Kitti2015. We manually marked the outline of the objects for better illustration.}
    \label{Qualitative comparisons on KITTI2015}
 \end{figure}

%% file: src/figandtable/qualitative_comp1.tex
\begin{figure}[h]
    \centering
    \begin{minipage}[t]{0.16\textwidth}
    \centering
    \includegraphics[width=0.97\textwidth]{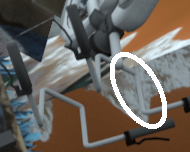}\\
    \includegraphics[width=0.97\textwidth]{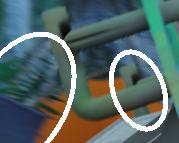}
    Color Image
    \end{minipage}
    \begin{minipage}[t]{0.16\textwidth}
    \centering
    \includegraphics[width=0.97\textwidth]{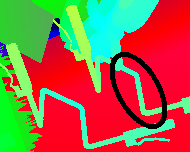}\\
    \includegraphics[width=0.97\textwidth]{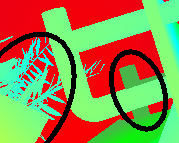}
    Ground truth
    \end{minipage}
    \begin{minipage}[t]{0.16\textwidth}
    \centering
    \includegraphics[width=0.97\textwidth]{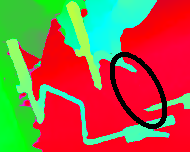}\\
    \includegraphics[width=0.97\textwidth]{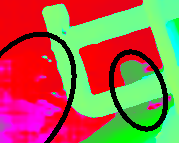}
    PSMNet
    \end{minipage}
    \begin{minipage}[t]{0.16\textwidth}
    \centering
    \includegraphics[width=0.97\textwidth]{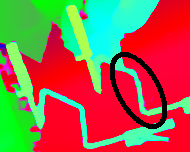}\\
    \includegraphics[width=0.97\textwidth]{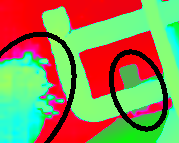}
    SG-PSMNet
    \end{minipage}
    \begin{minipage}[t]{0.16\textwidth}
    \centering
    \includegraphics[width=0.97\textwidth]{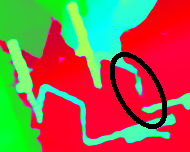}\\
    \includegraphics[width=0.97\textwidth]{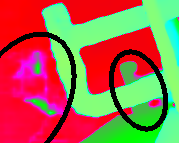}
    MSN2D
    \end{minipage}
    \begin{minipage}[t]{0.16\textwidth}
    \centering
    \includegraphics[width=0.97\textwidth]{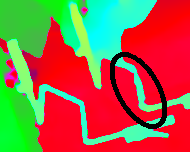}\\
    \includegraphics[width=0.97\textwidth]{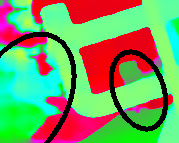}
    SG-MSN2D
    \end{minipage}
    \caption{Qualitative comparisons on Sceneflow.}
    \label{Qualitative comparisons on SceneFlow}
 \end{figure}

%% file: src/figandtable/app_fig.tex
\begin{figure}[h]
    \centering
    
    \begin{minipage}[t]{0.39\textwidth}
        \centering
    \includegraphics[width=0.97\textwidth]{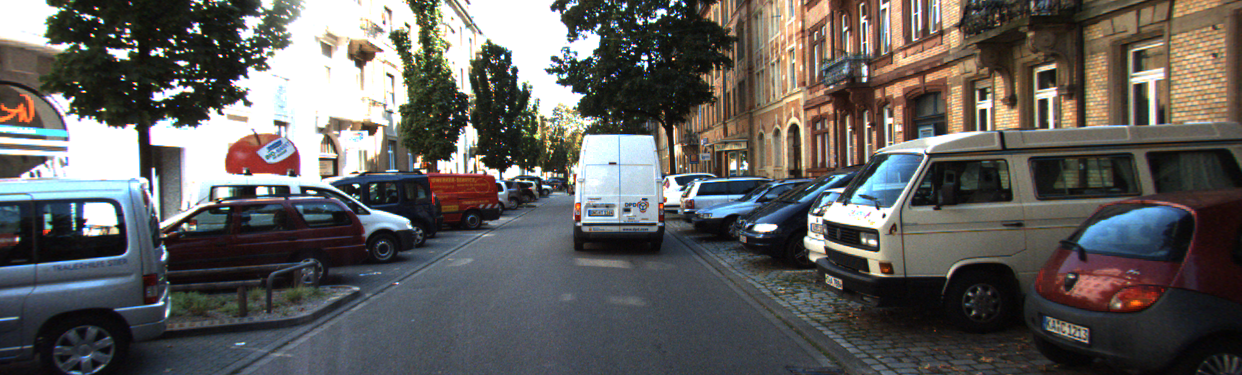}
    \includegraphics[width=0.97\textwidth]{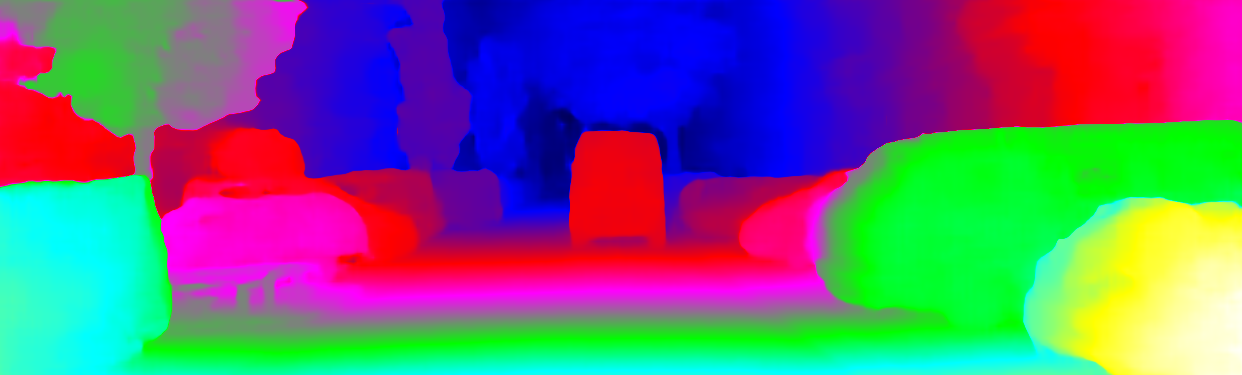}
    \includegraphics[width=0.97\textwidth]{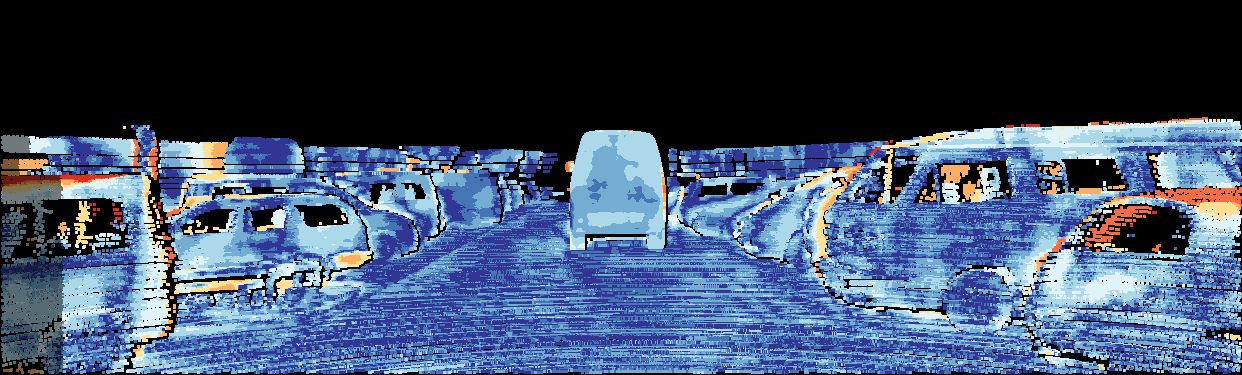}
    PSMNet
    \end{minipage}
    \begin{minipage}[t]{0.39\textwidth}
    \centering    
    \includegraphics[width=0.97\textwidth,height=0.0\textwidth]{fig/app_fig/000005_10.png}
    \includegraphics[width=0.97\textwidth]{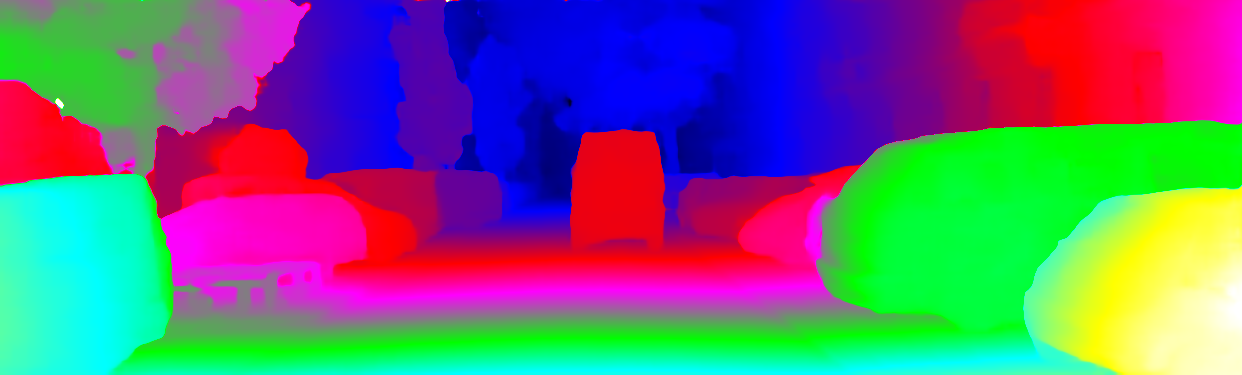}
    \includegraphics[width=0.97\textwidth]{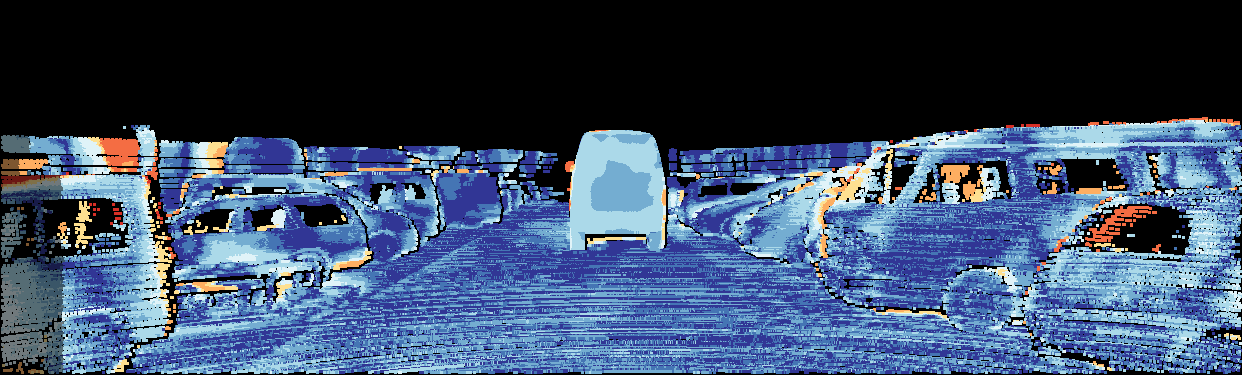}
    SG-PSMNet
    \end{minipage}

    \caption{ALL-D1$_{bg}$, ALL-D1$_{fg}$, ALL-D1$_{all}$ are PSMNet: ($3.67, 1.16, 3.45$), SG-PSMNet: ($3.24, 1.49, 3.08$)
    }
 \end{figure}

 \begin{figure}[h]
    \centering
    
    \begin{minipage}[t]{0.39\textwidth}
        \centering
    \includegraphics[width=0.97\textwidth]{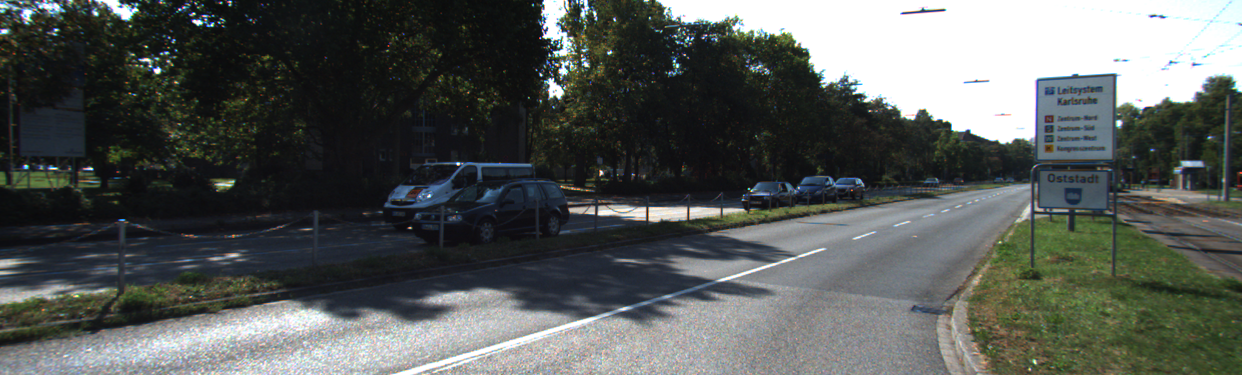}
    \includegraphics[width=0.97\textwidth]{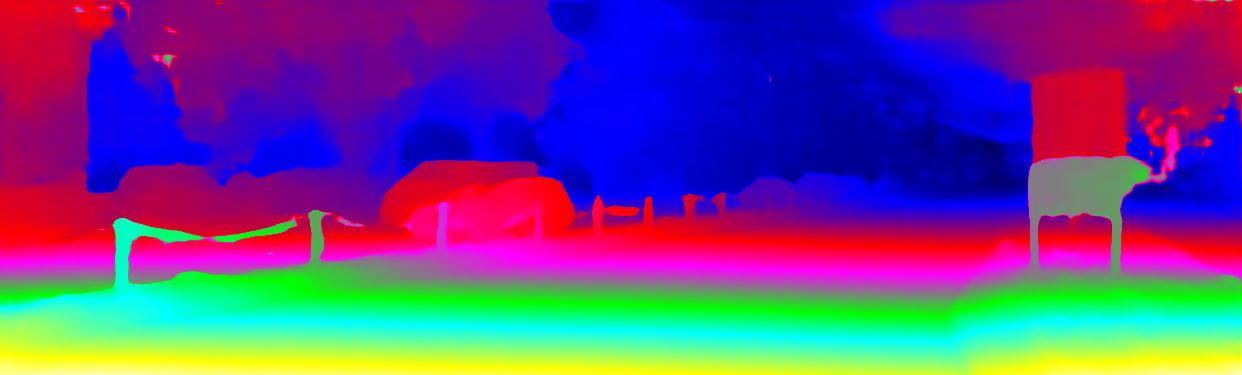}
    \includegraphics[width=0.97\textwidth]{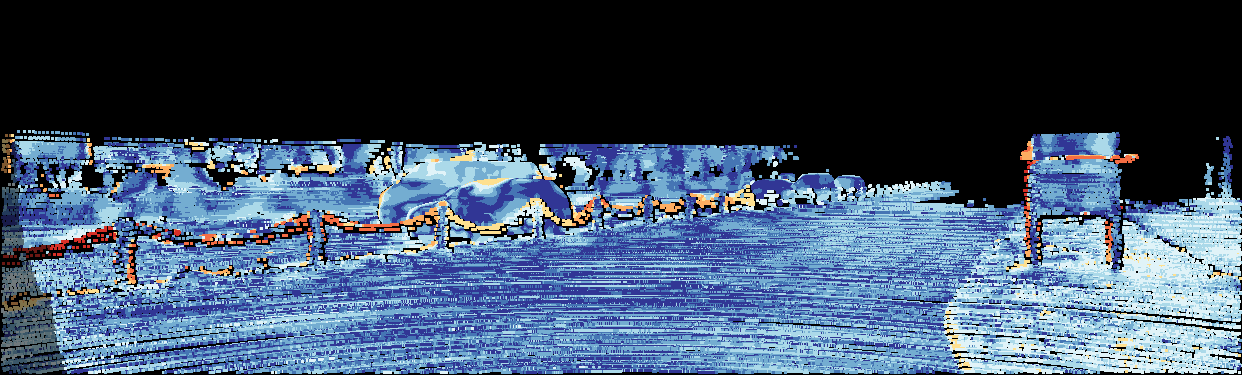}
    PSMNet
    \end{minipage}
    \begin{minipage}[t]{0.39\textwidth}
    \centering    
    \includegraphics[width=0.97\textwidth,height=0.0\textwidth]{fig/app_fig/000001_10.png}
    \includegraphics[width=0.97\textwidth]{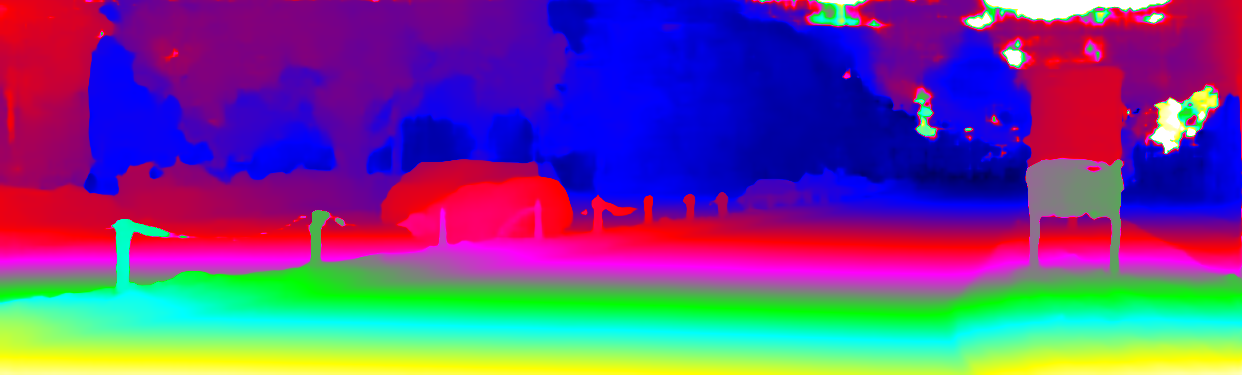}
    \includegraphics[width=0.97\textwidth]{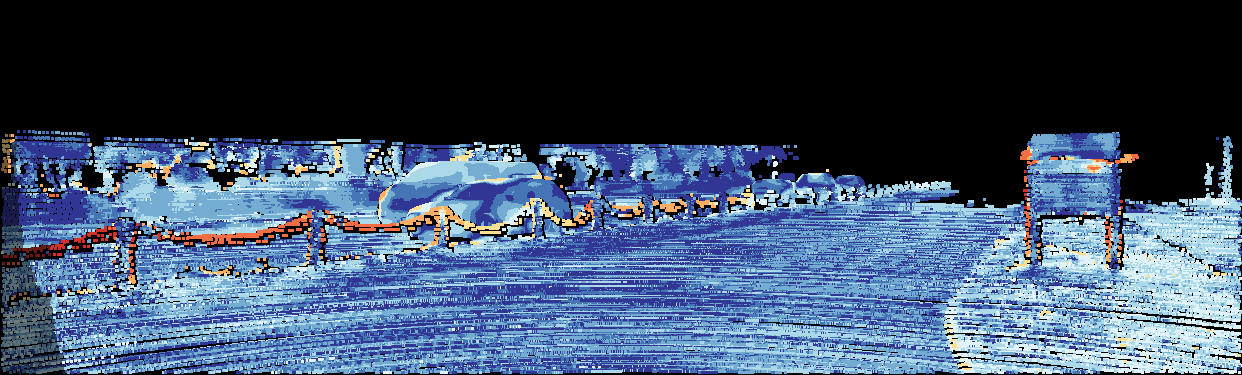}
    SG-PSMNet
    \end{minipage}

    \caption{ALL-D1$_{bg}$, ALL-D1$_{fg}$, ALL-D1$_{all}$ are PSMNet: ($1.96, 2.22, 1.99$), SG-PSMNet: ($1.66, 0.93,1.58$)}
 \end{figure}
